\documentclass{article} 
\usepackage{arxiv}
\renewcommand{\headeright}{}
\renewcommand{\undertitle}{}
\renewcommand{\headrulewidth}{0pt}
\usepackage{natbib}
\usepackage[T1]{fontenc}   

\usepackage{amsmath,amsfonts,bm}

\def\eqref#1{equation~\ref{#1}}

\def\1{\bm{1}}

\DeclareMathAlphabet{\mathsfit}{\encodingdefault}{\sfdefault}{m}{sl}
\SetMathAlphabet{\mathsfit}{bold}{\encodingdefault}{\sfdefault}{bx}{n}

\usepackage{hyperref}
\usepackage{url}
\usepackage{graphicx}
\usepackage{tikz}
\usetikzlibrary{arrows.meta, positioning, calc, fit, backgrounds}
\usepackage{booktabs}
\usepackage{amsmath}
\usepackage{amssymb}
\usepackage{xcolor}
\usepackage{multirow}
\usepackage{tabularray}
\usepackage{subcaption}
\usepackage[section]{placeins}

\newcommand{\benchname}{MeasureVQA} 

\title{When Do Cheap Probes Predict Expensive Training?\\
Probing 3D-CT Encoders for Text Generation}

\author{
\textbf{Renjie Liang\textsuperscript{1}, Zijian Xu\textsuperscript{1}, Jinqian Pan\textsuperscript{1}, Chengkun Sun\textsuperscript{1}, Zhengkang Fan\textsuperscript{1},} \\
\textbf{Shawn Li\textsuperscript{2}, You Qin\textsuperscript{3}, Mei Liu\textsuperscript{1}, Jie Xu\textsuperscript{1}} \\[4pt]
\small\textsuperscript{1}University of Florida, Gainesville, FL, USA \\
\small\textsuperscript{2}University of Southern California, Los Angeles, CA, USA \\
\small\textsuperscript{3}National University of Singapore, Singapore
}
\date{}

\begin{document}

\maketitle

\begin{abstract}
Building a 3D~CT vision language model begins with a choice of which image encoder to build on. Today that choice is made by fine-tuning every candidate through the full language model and comparing downstream scores, an enormously expensive search. A cheap probe on the encoder's representation promises a way out, but whether it forecasts the expensive outcome has never been tested. We test this with \textbf{CheapCT} on report generation and on \benchname{}, a new VQA dataset we build. \benchname{} scores the outcome one capability at a time, its answers measured from segmentation masks and Hounsfield units. Report generation scores the whole report at once and reflects mostly disease. CheapCT forecasts expensive training across every capability. The forecast survives changing the probe readout and the language-model backbone. The rank agreement between CheapCT and fine-tuning stays high throughout, from $\rho=0.90$ to $0.97$. Used to choose an encoder, CheapCT picks one nearly as good as the best while fine-tuning a single candidate, at orders of magnitude less compute. We release the code and \benchname{} at \url{https://github.com/renjie-liang/CheapCT}.

\end{abstract}

\section{Introduction}
\label{sec:intro}
Computed tomography is a workhorse of modern diagnostic imaging, with about $93$ million examinations performed in the United States in 2023~\citep{smithbindman2025ct}, and a single chest CT is a dense 3D volume that a radiologist reads slice by slice for many distinct findings. Automating that reading with a vision language model would place a second reader in every clinic, and the past two years have produced a steady stream of 3D~CT models that write reports and answer questions about a scan~\citep{10.1007/978-3-031-72390-2_45,hamamci2026generalist,bai2024m3dadvancing3dmedical,wu2023radfm,blankemeier2024merlin,11145341}. These models share one recipe, the visual instruction tuning introduced for natural images~\citep{NEURIPS2023_6dcf277e}. A pretrained 3D image encoder turns the volume into a set of visual tokens, a projector maps those tokens into a large language model, and the model is fine-tuned to produce text. The encoder strongly shapes the final quality, since the pretrained representation a model starts from is a major driver of how well it transfers to a downstream task~\citep{8954384,9577835,ICLR2024_47360925}. Today the encoder is chosen by fine-tuning every candidate through the full model and comparing downstream scores, a search costly enough to dominate the research cycle.
A cheap probe trains a lightweight readout on a representation to measure what it encodes~\citep{alain2018understandingintermediatelayersusing}. It offers a way to select a 3D~CT encoder without full vision-language fine-tuning. In neural architecture search, low-cost proxies predict how candidate networks rank after full training~\citep{Abdelfattah2021ZeroCostPF}. Whether a representation probe can similarly predict downstream performance after full fine-tuning in 3D~CT vision language models remains unexplored.

Report generation is a widely studied task~\citep{10.1007/978-3-031-72390-2_45}, and in 3D~CT vision language modeling it is the one that draws on a model's capabilities most comprehensively~\citep{WANG2025103627}. But a report is scored by a single number that reflects mostly disease, so against it alone a probe can be judged only in aggregate. The capabilities we mean are the five a radiologist reads separately: size, density, location, texture, and disease~\citep{aerts2014decoding}. To resolve the probe's forecast into these, we build \benchname{}, a dataset whose answers are measured from segmentation masks and Hounsfield units. Each question targets one capability, so a density probe is graded against a density question. We measure the probe against report generation and \benchname{} in parallel.

We propose \textbf{CheapCT}, a cheap probe that forecasts expensive training. We test it at scale, probing nine encoder variants on CT-RATE~\citep{hamamci2026generalist} and a range of readout configurations, each probe trained in minutes. We carry the same variants through full fine-tuning on four language-model backbones. We then extend the same test to a second corpus, Merlin~\citep{blankemeier2024merlin}. The forecast holds across a wide band of readouts, and CheapCT is the position-aware one, strong across the board and the only kind that recovers spatial location, which a position-blind probe misses. We make three contributions:
\begin{itemize}
\item We give the first evidence that a cheap probe on a pretrained encoder's representation can forecast the outcome of expensive 3D~CT vision language fine-tuning, one capability at a time, robust to the choice of probe readout and language-model backbone.
\item We build and release \benchname{}, a 3D~CT VQA whose answers are measured from segmentation masks and Hounsfield units and split into five clinical capabilities, each audited end to end.
\item Used as an encoder-selection rule, CheapCT (choosing an encoder by its probe score) recovers most of what fine-tuning every candidate would find at a fraction of the cost, with controls that keep the claim falsifiable.
\end{itemize}

\begin{figure}[t]
\centering
\includegraphics[width=\linewidth]{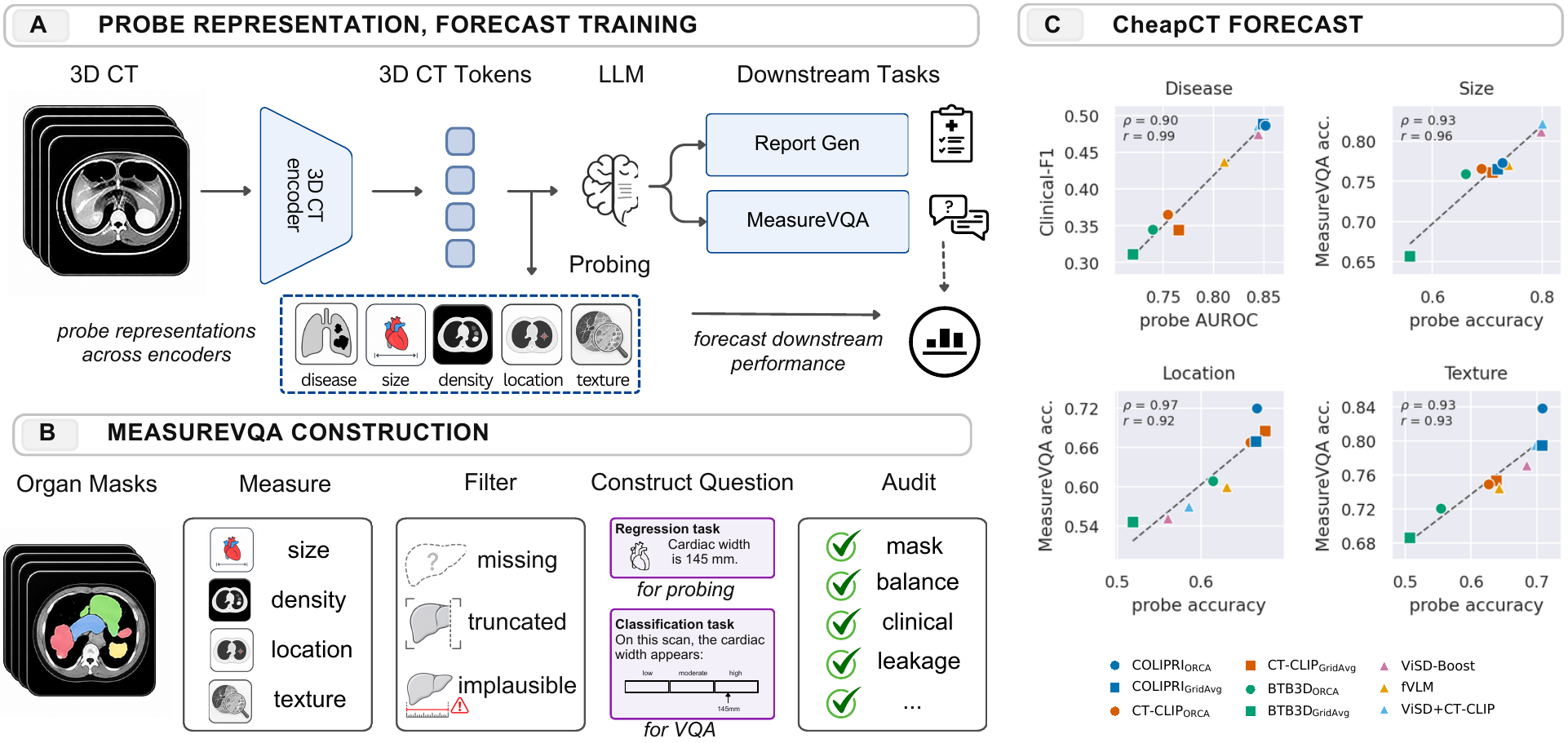}
\caption{\textbf{(A)}~A language model is fine-tuned on 3D-CT encoder tokens. A cheap probe reads the same representation and forecasts that outcome without the fine-tuning. \textbf{(B)}~\benchname{} construction, from organ masks and Hounsfield units. \textbf{(C)}~Per capability, the probe score against the downstream score it forecasts across the encoders, on CT-RATE with Llama-3.1-8B. The probe \textbf{CheapCT} forecasts the expensive outcome.}
\label{fig:overview}
\end{figure}

\section{Related Work}
\label{sec:related}
\noindent\textbf{3D-CT vision language models.} 3D-CT vision language models are built on a pretrained image encoder, and encoders now span every pretraining paradigm, self-supervised~\citep{zhou2021modelsgenesis,10658544,pai2025visionfoundationmodelscomputed}, supervised~\citep{ICLR2024_47360925}, and contrastive image-text~\citep{hamamci2026generalist,blankemeier2024merlin}. A growing set of models fine-tune such an encoder to write reports and answer questions~\citep{wu2023radfm,bai2024m3dadvancing3dmedical,10.1007/978-3-031-72390-2_45,11145341}, so which encoder to build on is a consequential choice, yet an encoder's downstream value can be judged today only by fine-tuning a full model on it. A separate line grounds these outputs in anatomy, as in RadGenome-Chest~CT~\citep{zhang2024radgenomechestctgroundedvisionlanguage}, whose region labels come from reports. However, we measure our answers from the images instead, so each answer is a physical measurement rather than a report annotation.

\noindent\textbf{Representation probing.} A linear or attention probe reads a task off a representation and is a standard, inexpensive gauge of what that representation encodes~\citep{alain2018understandingintermediatelayersusing}. For model selection, probe and upstream accuracy correlate with full fine-tuning transfer~\citep{8954384}, though the correlation weakens for self-supervised models and off-distribution tasks~\citep{9577835,baharoon2024evaluatinggeneralpurposevision}. A related line forecasts transfer without any fine-tuning, scoring a model from its features alone~\citep{pmlr-v119-nguyen20b,pmlr-v139-you21b}. We bring this idea to 3D-CT text generation and, unlike prior work, test when a cheap probe forecasts training one capability at a time rather than assume it transfers.

\noindent\textbf{Performance prediction in AutoML.} Low-cost proxies rank candidate networks in a single pass, without training each one~\citep{Abdelfattah2021ZeroCostPF,ea2d8087f87e4153a0e2941651552995}, and scaling studies predict a large model's behavior from a small proxy run. Closest to us, probing predicts fine-tuning performance in NLP~\citep{zhu-etal-2022-predicting}. We bring this proxy role to encoder selection for 3D-CT vision language models, and add a capability-resolved test that survives a change of readout and of backbone.
\section{The \benchname{} Dataset}
\label{sec:benchmark}

\subsection{Design principles}
\label{sec:design}
A generated report collapses many kinds of visual information into one aggregate score that reflects mostly disease, so it cannot separately show how well an encoder captures finer, capability-level information. We organize this information into five capabilities that a radiologist assesses separately. They are size, density, location, texture, and disease. We build a dataset that scores each on its own. We define size through organ volumes, diameters, and proportions, density through tissue attenuation, location through relative spatial relationships, and texture through the shape of intensity distributions. Within each capability, guided by the quantitative-CT literature and clinical measurement standards, we select attributes that are reliably measurable, clinically meaningful, and well separated from one another. All measurements are computed from organ masks and the Hounsfield values within them, not from reports or a language model. Disease labels, in contrast, come from the reports. These clinically distinct but related dimensions organize the attributes and questions of the dataset.

\begin{figure}[tbh]
\centering
\subcaptionbox{Measured from masks}{\includegraphics[height=4.3cm]{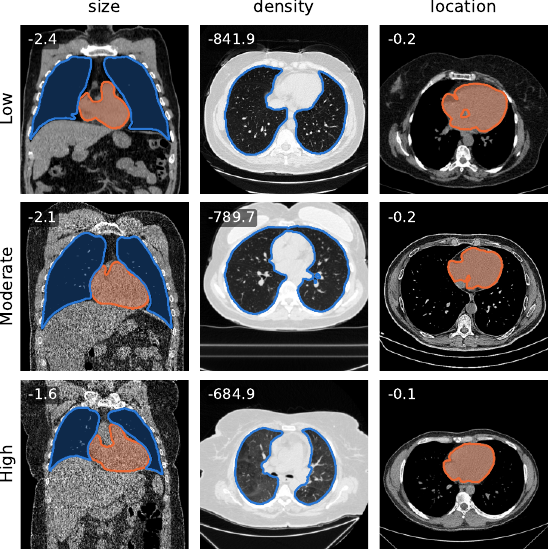}}\hfill
\subcaptionbox{Balanced classes}{\includegraphics[height=4.3cm]{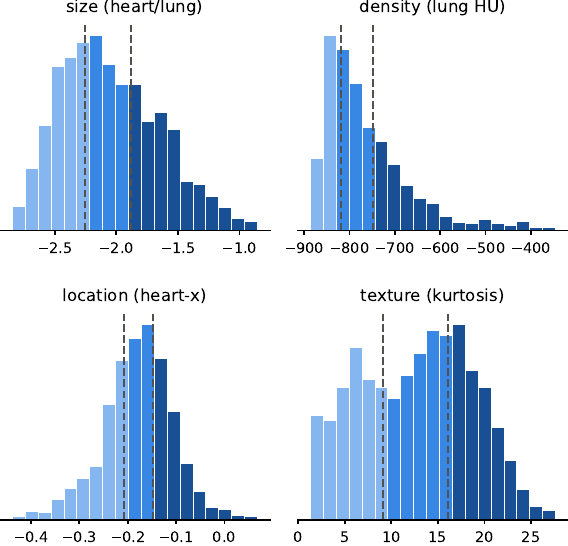}}\hfill
\subcaptionbox{Clinical validity}{\includegraphics[height=4.3cm]{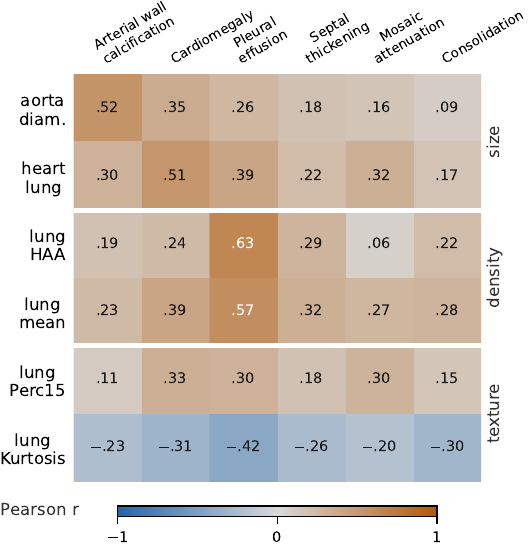}}
\caption{\textbf{(a)}~CT-RATE slices per class, with the organ mask and the measured value. \textbf{(b)}~Per-attribute value distributions by class, dashed lines the frozen tertile cutpoints. \textbf{(c)}~Signed Pearson $r$ between each measurement and the abnormalities.}
\label{fig:dataset}
\end{figure}

\subsection{Dataset construction}
\label{sec:corpora}
\benchname{} is constructed from two public 3D~CT datasets. CT-RATE provides non-contrast chest CT and Merlin contrast-enhanced abdominal CT. The two sources cover different body regions and acquisition conditions, allowing the same types of clinical capability to be constructed in the chest and abdomen. We follow their existing training and validation splits. Figure~\ref{fig:overview}(B) sketches the construction pipeline, from organ masks to audited questions. For each CT volume we use TotalSegmentator~\citep{wasserthal2023totalsegmentator} organ masks, from which we extract continuous measurements using mask geometry and the Hounsfield values within the masks. These include organ volumes and ratios, anatomical diameters, normalized positions, tissue attenuation, intensity percentiles, and distribution shape. We then filter unreliable values caused by missing, truncated, or erroneous segmentations using organ presence, physical mask size, Hounsfield composition, and anatomical plausibility. These rules use neither disease labels nor the extremity of a value in the population distribution. Filtering is applied per attribute, so an unreliable organ removes only its own measurements rather than the whole volume. The resulting filtered measurements form the basis for the audit and the questions. Figure~\ref{fig:dataset}(a) shows a representative slice per class, the value read off the mask.

\subsection{Dataset audit}
\label{sec:gates}
We audit the dataset with statistical and analytic checks on the masks, measurements, and reports, independent of the probes and encoders we evaluate. The masks are reliable. Each organ is present in about $98\%$ of CT-RATE volumes and over $98\%$ of Merlin volumes, and guards withhold rather than corrupt the rare failures. Repeat measurements agree closely, for example a test-retest of $0.969$ for vertebral bone density. Figure~\ref{fig:dataset}(b) plots the value distributions against their frozen cutpoints. The classes are balanced, every tertile holding between $0.30$ and $0.39$ of the mass and the reference letter uniform to within a third of a percentage point, so a baseline that always answers the most frequent letter scores at chance.

Across capabilities, each measurement's extreme decile is enriched for the disease it implies. Cardiomegaly shows an $11\times$ lift and splenomegaly a $70\times$ lift. This is convergent evidence from the masks and the reports. Figure~\ref{fig:dataset}(c) shows this per measurement, texture the exception because disease flattens its intensity histogram. For shortcuts, a size proxy built from raw organ voxel count stays near its baseline on most density, location, and texture attributes, so organ scale alone does not explain them, and the few measurable couplings are flagged. The full audit, covering mask quality, thresholds and balance, and shortcuts, is in Appendix~\ref{app:audit}.

\begin{table}[!htb]
\centering
\scriptsize
\setlength{\tabcolsep}{3pt}
\renewcommand{\arraystretch}{1.06}
\caption{\benchname{} statistics by corpus and capability. Reliability is the reproducibility of a measurement across reconstructions of the same scan. Question counts combine training and validation and exclude open-value questions. Chance is the random-guessing accuracy.}
\label{tab:measurevqa-stats}
\begin{tabular}{@{}llrrccrcc@{}}
\toprule
\textbf{Corpus} & \textbf{Capability} & \textbf{Attributes} & \textbf{Volumes} & \textbf{Coverage\,(\%)} & \textbf{Reliability} & \textbf{Questions} & \textbf{Avg.\,words} & \textbf{Chance} \\
\midrule
\multirow{5}{*}{CT-RATE}
 & Size     & 5  & 49{,}371 & 98.4  & 0.965 & 246{,}504 & 9.0  & 0.367 \\
 & Density  & 4  & 49{,}411 & 98.5  & 0.956 & 195{,}892 & 9.8  & 0.374 \\
 & Location & 3  & 49{,}382 & 98.4  & 0.949 & 148{,}001 & 10.0 & 0.333 \\
 & Texture  & 3  & 43{,}359 & 86.4  & 0.592 & 130{,}075 & 11.0 & 0.389 \\
 & Disease  & 18 & 50{,}188 & 100.0 & ---   & 903{,}384 & 7.5  & 0.500 \\
\midrule
\multirow{4}{*}{Merlin}
 & Size     & 3  & 20{,}356 & 100.0 & --- & 60{,}060  & 8.7  & 0.444 \\
 & Density  & 3  & 20{,}360 & 100.0 & --- & 60{,}409  & 9.0  & 0.388 \\
 & Location & 2  & 19{,}977 & 98.1  & --- & 39{,}954  & 13.0 & 0.333 \\
 & Disease  & 30 & 20{,}189 & 99.2  & --- & 136{,}052 & 7.1  & 0.500 \\
\midrule
\multicolumn{2}{@{}l}{\textbf{Total}} & \textbf{71} & \textbf{70{,}548} & & & \textbf{1{,}920{,}331} & & \\
\bottomrule
\end{tabular}
\end{table}

\subsection{VQA and statistics}
\label{sec:two-tasks}
After auditing, each measurement is released in two forms that share one underlying value, an open-value question that requests the exact number and a discretized letter question. The letter question is built by cutting most measurements into low, moderate, and high tertiles and a few into two classes, a clinical threshold replacing the tertile only where the measured quantity matches the guideline and its split stays balanced enough. The cutpoints are frozen on the training split and applied unchanged to validation, and the label is then rendered as a multiple-choice question from a fixed template with options shuffled so answer position carries no signal. Disease is asked instead as a presence question with report-derived labels. Because both forms come from the same number, a question and its value are matched by construction. 
\benchname{} draws on $47{,}149$ training and $3{,}039$ validation volumes from CT-RATE and $15{,}306$ and $5{,}054$ from Merlin, posing about $1.92$M letter-scored questions and a further $0.45$M open-value questions, summarised in Table~\ref{tab:measurevqa-stats}.

\section{Method}
\label{sec:method}
\subsection{The representation probe}
\label{sec:probe}
We propose a lightweight probe that measures how much of a capability an encoder's representation carries. The probe takes the same visual tokens the language model would receive, pools them into a single vector, and predicts that capability's attributes, a regression for the continuous measurements and a classification for disease. It never trains the language model. A small readout, a linear pool, a gated attention, or a small transformer, trains in minutes on a single GPU. We vary the pooling mechanism, its capacity, and the input normalization across a family of readouts. The headline, which we call CheapCT, is a position-aware attention pool and the strongest all-round reader in the family. Each probe reports one score per capability on the validation split. The full readout family and its training protocol are in Appendix~\ref{app:probe}.

\subsection{The testbed}
\label{sec:testbed}
\textbf{Encoders.} The encoder is the axis we vary. We compare six pretrained 3D-CT encoders on CT-RATE, spanning the pretraining paradigms in use today. They are the contrastive image-text encoders COLIPRI~\citep{wald2026comprehensivelanguageimagepretraining3d} and CT-CLIP~\citep{hamamci2026generalist}, the discrete visual tokenizer BTB3D~\citep{NEURIPS2025_c4861cc8}, and the organ-aware encoders ViSD~\citep{11444916}, fVLM~\citep{ICLR2025_3c3b120a}, and a combination of ViSD and CT-CLIP~\citep{liang2026embeddingbottleneckadaptiveretrievalaugmented}. A dense grid carries more tokens than a language model can accept, so we reduce it to a common budget by one of two standard aggregations, an organ-centroid pooling (ORCA)~\citep{liang2026orcaorgancentroidaggregationtrainingfree} or a plain grid average (GridAvg), and take each (encoder, aggregation) pair as a representation of its own. Encoders that already emit organ-level tokens enter directly. This yields nine representation variants, the points our correlations run over.

\textbf{Downstream tasks.} The expensive outcome is a language model fine-tuned on an encoder to produce text. Report generation draws on every capability at once but scores mostly disease, and \benchname{} supplies the four measurement capabilities one question at a time, so the two image-grounded tasks together expose all five.

\textbf{Backbones.} Because the outcome could depend on the language model as much as on the encoder, we run four backbones, Llama-3.1-8B, the larger Llama-3.3-70B, Gemma-3-27B, and its medically pretrained sibling MedGemma-27B, covering different model sizes and degrees of domain specialization. Each encoder is attached to a backbone by the same lightweight projector and adapter, trained in stages and held fixed so that from one encoder to the next only the encoder changes. Every encoder is carried through full fine-tuning on every backbone, so each backbone is an independent test of whether the one probe score forecasts training.
\subsection{Measuring forecast agreement}
\label{sec:validity}
For a single encoder, each capability gives a probe score, AUROC for disease and $R^2$ (variance explained) for the four continuous measurements, set against the downstream score for the same capability. Disease is scored on report generation by CE-F1 (clinical-efficacy F1)~\citep{yan2022radbert} and GREEN~\citep{ostmeier-etal-2024-green}, and each measurement capability by the accuracy of its \benchname{} questions. The probe predicts the very value each measurement question is built from, so the two are matched by construction.

Forecast agreement comes from varying the encoder. Holding a capability fixed, each encoder gives a probe score and a paired downstream score, and we correlate the two across the encoders. We summarise each pairing with three standard correlation coefficients, all running from $-1$ to $1$. Spearman $\rho$~\citep{85f53594-1421-3dd6-8318-2f0686ede307} is the rank correlation, high when the probe orders the encoders as training does. Pearson $r$~\citep{a0dc553c-0830-3fe2-ab2a-eece0d66a7db} is the raw-score correlation, high when probe and downstream scores line up tightly enough to forecast from. Weighted Kendall $\tau$~\citep{10.1145/2736277.2741088} is a rank correlation that weights the top of the order, the region model selection cares about, since a practitioner fine-tunes only the encoders the probe ranks first. Where these are high the cheap probe stands in for the expensive run. We count a capability as forecastable only where this holds on each of the four backbones, whose downstream scores the probe never sees, and survives the controls that rule out artifacts.
\section{Experiments}
\label{sec:experiments}

\subsection{Forecast agreement}
\label{sec:validity-results}

Table~\ref{tab:main_results} reports the downstream score each of the nine encoders reaches after fine-tuning on CT-RATE with Llama-3.1-8B, beside its CheapCT score. For each capability, we correlate the two across the encoders as a rank and a linear agreement, Spearman $\rho$ and Pearson $r$, plotted in Figure~\ref{fig:overview}. Training curves are in Appendix~\ref{app:training}.

\begin{table*}[!t]
\centering
\scriptsize
\setlength{\tabcolsep}{2pt}
\begin{tabular}{@{}l ccccc cc cc cc cc@{}}
\toprule
& \multicolumn{5}{c}{Report generation} & \multicolumn{8}{c}{\benchname{}} \\
\cmidrule(lr){2-6}\cmidrule(lr){7-14}
& \multicolumn{5}{c}{Disease} & \multicolumn{2}{c}{Size} & \multicolumn{2}{c}{Density} & \multicolumn{2}{c}{Location} & \multicolumn{2}{c}{Texture} \\
\cmidrule(lr){2-6}\cmidrule(lr){7-8}\cmidrule(lr){9-10}\cmidrule(lr){11-12}\cmidrule(lr){13-14}
Encoder & CE-F1 & GREEN & BLEU-4 & ROUGE & CheapCT & LLM & CheapCT & LLM & CheapCT & LLM & CheapCT & LLM & CheapCT \\
\midrule
\textit{Noise floor}   & $0.20$ & $0.12$ & $0.08$ & $0.18$ & -- & $0.45$ & -- & $0.40$ & -- & $0.33$ & -- & $0.52$ & -- \\
\textit{Shuffle floor} & $0.28$ & $0.29$ & $0.18$ & $0.29$ & -- & $0.45$ & -- & $0.38$ & -- & $0.33$ & -- & $0.50$ & -- \\
\midrule
BTB3D$_\mathrm{GridAvg}$   & $0.31$ & $0.25$ & $0.19$ & $0.31$ & $0.72$ & $0.66$ & $0.56$ & $0.65$ & $0.52$ & $0.55$ & $0.52$ & $0.69$ & $0.51$ \\
BTB3D$_\mathrm{ORCA}$   & $0.34$ & $0.30$ & $0.20$ & $0.31$ & $0.74$ & $0.76$ & $0.66$ & $0.70$ & $0.53$ & $0.61$ & $0.61$ & $0.72$ & $0.55$ \\
COLIPRI$_\mathrm{GridAvg}$ & $0.49$ & $0.33$ & $0.21$ & $0.32$ & $0.85$ & $0.77$ & $0.72$ & $0.81$ & $0.74$ & $0.67$ & $0.67$ & $0.79$ & $0.71$ \\
COLIPRI$_\mathrm{ORCA}$ & $0.49$ & $0.35$ & $0.20$ & $0.32$ & $0.85$ & $0.77$ & $0.73$ & $0.84$ & $0.75$ & $0.72$ & $0.67$ & $0.84$ & $0.71$ \\
CT-CLIP$_\mathrm{GridAvg}$ & $0.34$ & $0.31$ & $0.20$ & $0.32$ & $0.77$ & $0.76$ & $0.71$ & $0.73$ & $0.58$ & $0.68$ & $0.68$ & $0.75$ & $0.64$ \\
CT-CLIP$_\mathrm{ORCA}$ & $0.37$ & $0.30$ & $0.19$ & $0.31$ & $0.75$ & $0.77$ & $0.69$ & $0.71$ & $0.57$ & $0.67$ & $0.66$ & $0.75$ & $0.63$ \\
fVLM                    & $0.44$ & $0.32$ & $0.21$ & $0.32$ & $0.81$ & $0.77$ & $0.74$ & $0.72$ & $0.67$ & $0.60$ & $0.63$ & $0.74$ & $0.64$ \\
ViSD-Boost              & $0.47$ & $0.31$ & $0.21$ & $0.32$ & $0.84$ & $0.81$ & $0.80$ & $0.78$ & $0.71$ & $0.55$ & $0.56$ & $0.77$ & $0.68$ \\
ViSD+CT-CLIP            & $0.49$ & $0.32$ & $0.21$ & $0.33$ & $0.84$ & $0.82$ & $0.80$ & $0.81$ & $0.70$ & $0.57$ & $0.59$ & $0.79$ & $0.70$ \\
\midrule
Spearman $\rho$        & $0.90$ & $0.97$ & $0.62$ & $0.60$ & -- & $0.93$ & -- & $0.93$ & -- & $0.97$ & -- & $0.93$ & -- \\
Pearson $r$            & $0.99$ & $0.83$ & $0.73$ & $0.76$ & -- & $0.96$ & -- & $0.91$ & -- & $0.92$ & -- & $0.93$ & -- \\
Weighted Kendall $\tau$ & $0.73$ & $0.94$ & $0.32$ & $0.32$ & -- & $0.89$ & -- & $0.84$ & -- & $0.83$ & -- & $0.87$ & -- \\
\bottomrule
\end{tabular}
\caption{The downstream LLM score beside the CheapCT probe score on CT-RATE with Llama-3.1-8B. The CheapCT probe score is AUROC for disease and tertile accuracy for the measurements. The two \textit{floor} rows are averaged over the encoders. CheapCT has no floor.}
\label{tab:main_results}
\end{table*}

\noindent\textbf{Report generation.} CE-F1 separates the encoders widely. CheapCT tracks it tightly, at $\rho=0.90$ and $r=0.99$, the top three encoders falling within $0.003$, which holds $\rho$ just below $r$. GREEN judges the whole report, well beyond the eighteen disease labels, yet CheapCT forecasts it about as tightly as CE-F1, at $\rho=0.97$ and $r=0.83$. The surface $n$-gram metrics, BLEU-4 and ROUGE-L, score word overlap instead. They stay nearly flat across the encoders, so they carry no forecast and measure fluency rather than clinical correctness.

\noindent\textbf{\benchname{}.} Size, density, and texture are forecast at $\rho=0.93$, and location, the spatial capability, at $\rho=0.97$. The answer is a single letter, so the accuracy measures the capability directly, with little generation noise between the encoder and the score. CheapCT predicts the very measurement each question is built from, so the probe and downstream share one target.

\noindent\textbf{Selection quality.} We also frame the forecast in the terms model selection uses. The Weighted Kendall $\tau$ in Table~\ref{tab:main_results} stays high on the measurements and dips on disease, where the leading encoders tie, and a permutation test rejects chance at $p<0.001$, with a wide interval at nine encoders. Appendix~\ref{app:credibility} gives the credibility checks. Choosing the encoder CheapCT ranks first recovers $95$ to $100\%$ of the best downstream score, the exact best on three of the five capabilities. CheapCT, trained in minutes, thus forecasts the expensive outcome well enough to select an encoder with.

\subsection{Across backbones}
\label{sec:backbone}
\begin{figure}[!tbh]
\centering
\includegraphics[width=\linewidth]{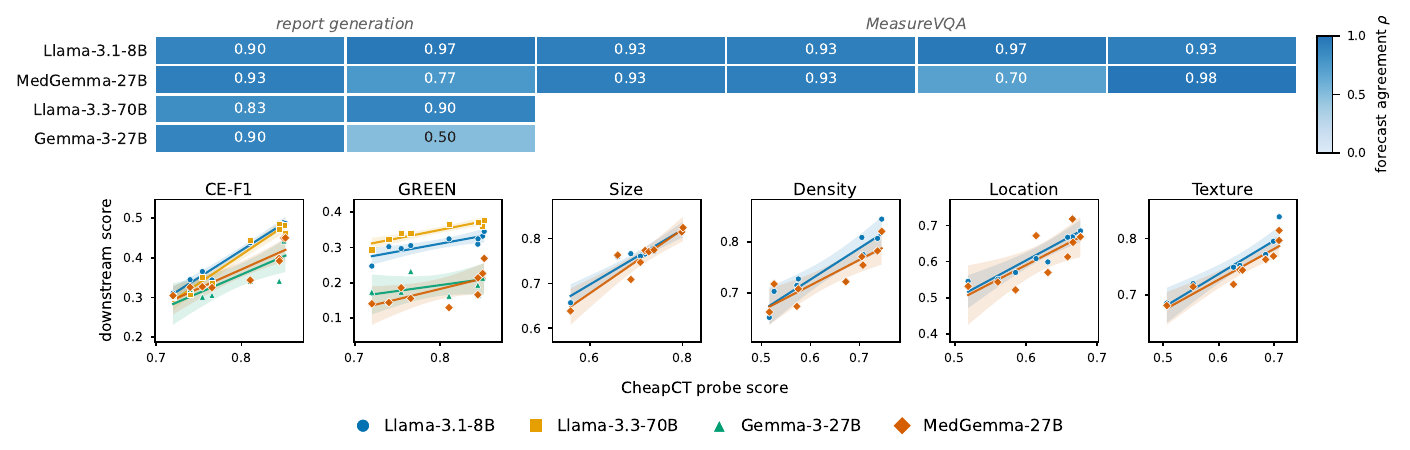}
\caption{Cross-backbone forecast agreement on CT-RATE. Top, $\rho$ per backbone and target. Bottom, CheapCT against the downstream score per capability, one point per encoder.}
\label{fig:backbone}
\end{figure}

CheapCT forecasts fine-tuning across language-model backbones. We carry the same nine encoders through full fine-tuning on four backbones and evaluate the forecast. Figure~\ref{fig:backbone} gives the cross-encoder $\rho$ per backbone. On report generation CheapCT tracks CE-F1 on all four backbones, at $\rho=0.83$ to $0.93$. On \benchname{} it tracks the measurements on both backbones we fine-tune it for, from $\rho=0.70$ on location under MedGemma to $\rho=0.98$ on texture.

The backbones differ in what they do best. Llama-3.1-8B is strongest on CE-F1, marginally over Llama-3.3-70B, and the two Gemma models trail. Llama-3.3-70B is the clear best on GREEN. The split fits their sizes. The small model already reads disease, which CE-F1 scores. The large model's broader capability shows on GREEN, which judges the whole report. CheapCT tracks GREEN on three of the four backbones, from $\rho=0.77$ to $0.97$. On Gemma-3-27B it is weaker, at $\rho=0.50$. Gemma is a general instruct model, and its reports arrive with conversational preambles and non-radiology formatting. We regenerate them with an explicit style instruction before scoring. We attribute this to its instruct prior leaking into the output, strong enough to shape the report in ways the holistic GREEN judge penalizes, even though the clinical labels CE-F1 scores stay well ordered.

\subsection{Robustness to the readout}
\label{sec:sensitivity}

\begin{figure*}[h]
\centering
\begin{minipage}{0.9\linewidth}
\centering
\subcaptionbox{Readout robustness\label{fig:sensitivity}}{\includegraphics[height=3.5cm]{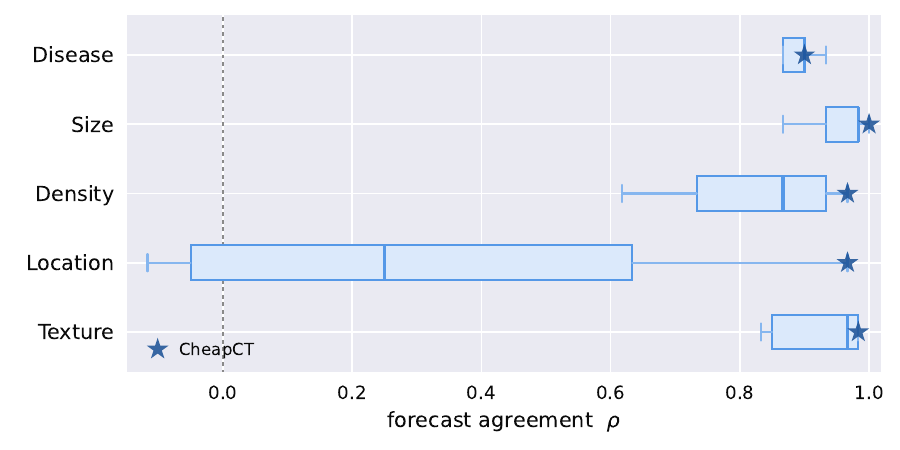}}\hfill
\subcaptionbox{Capability coupling\label{fig:coupling}}{\includegraphics[height=3.5cm]{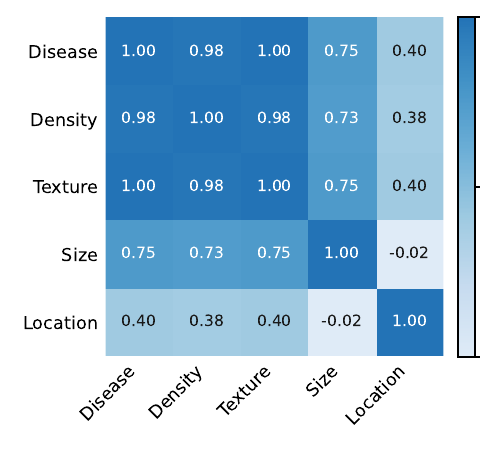}}
\end{minipage}
\caption{(a) For each capability, a box spanning the forecast agreement $\rho$ over the thirteen readouts. (b) The cross-encoder Spearman $\rho$ between the CheapCT probe scores of each capability pair, on the Llama-3.1-8B backbone.}
\end{figure*}

The forecast does not depend on the readout hyperparameters. We sweep the readout over a band of thirteen choices, varying its capacity, its pooling type, and the input normalization, shown in Figure~\ref{fig:sensitivity}. For four of the five capabilities the verdict does not move. Size, density, texture, and disease stay forecast across the band, their correlation never falling below $\rho=0.57$. For these the forecast is a property of the representation, not of the probe that reads it. Location is the exception, and an informative one. It is near $\rho=0$ for any permutation-invariant pool, a mean, a max, or a top-$k$, and it climbs to $\rho=0.96$ once the readout is position-aware. Location is a spatial relation, and the fine-tuned model can read it because it consumes the tokens as an ordered sequence. A permutation-invariant probe discards exactly that position signal, so it cannot forecast what the model does with location, while a position-aware probe keeps it and can. This is why CheapCT is the position-aware pool, the strongest all-round probe in the family and best or near-best on every capability. It also stays at a moderate capacity. The highest-capacity readouts, a four-layer transformer and the dual pool, extract less rather than more, the signature of a probe fitting the task instead of the representation.

\subsection{Capability coupling}
\label{sec:coupling}
An encoder that reads one capability well tends to capture the others well. The tendency is not uniform. Figure~\ref{fig:coupling} correlates the CheapCT probe scores across the nine encoders, one capability pair at a time. Disease, density, and texture move together almost perfectly, $\rho$ above $0.98$, and size tracks them at $\rho\approx0.75$. These four share one axis of encoder quality. Location does not join it. Its rank is near-independent of the rest, $\rho=0.40$ with disease and $-0.02$ with size, and a principal-components analysis agrees, one component for the shared axis and a second, about a fifth of the variance, for location alone. The downstream scores give the same two-axis structure.
This is a second sense in which location stands apart. A different pool is needed to extract it, and a different property of the encoder carries it, so a single quality score would rank it wrong. The forecast has to be made one capability at a time.

\subsection{The floor control}
\label{sec:floor}
A probe forecasts a downstream task only to the extent that the fine-tuned model itself reads the image. We re-score each trained model with its CT tokens replaced two ways. \emph{Noise} substitutes per-channel Gaussian tokens matched to the real tokens' mean and variance but without spatial structure. \emph{Shuffle} substitutes the real tokens of a randomly paired volume, a valid scan for the wrong patient. The score that survives is what the task solves without the image, and the gap to the real score is the image gain.

\begin{figure*}[!tbh]
\centering
\includegraphics[width=\linewidth]{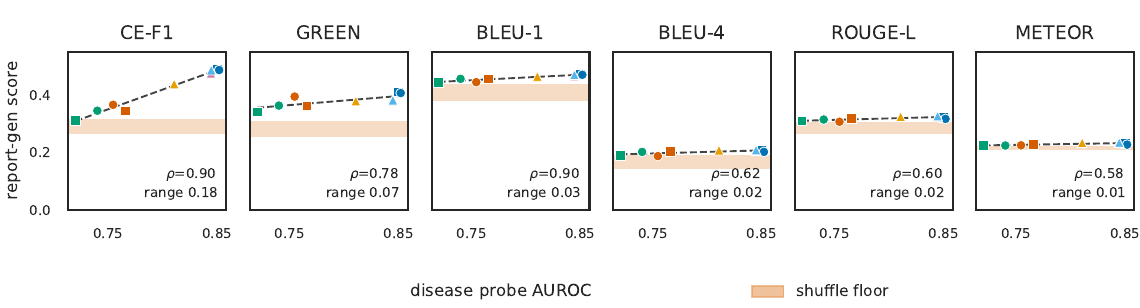}
\caption{Each report-generation metric against the CheapCT disease probe, on CT-RATE with Llama-3.1-8B.}
\label{fig:reportgen_fluency}
\end{figure*}

For report generation the metrics split. The clinical ones collapse. CE-F1 drops from $0.42$ with the real image to $0.20$ under noise and $0.28$ under shuffle, since the report is specific to the patient and cannot be produced without reading the scan, and the clinical judge GREEN moves the same way. This gap is an average, wide for the strong encoders and narrow for the weakest, whose report barely beats a mismatched scan. The surface $n$-gram metrics do not collapse. BLEU-4 and ROUGE-L score a report from a mismatched scan almost as high as the real one, and across the encoders they span two points where CE-F1 spans eighteen, plotted in Figure~\ref{fig:reportgen_fluency}. A metric that cannot tell a correct report from a plausible wrong one measures fluency, not clinical correctness. We treat these as a fluency floor and rank clinical quality by CE-F1, GREEN, and CheapCT.

\benchname{} collapses as sharply, each capability falling to the noise and shuffle floors in Table~\ref{tab:main_results}. Its question is identical across volumes, so the only thing separating a right answer from a wrong one is the measurement in the scan. The collapse is also a check on the benchmark. Shuffled options and image-only answers leave a mismatched volume no textual shortcut, and the fall to the floor confirms none was left open. Report generation and \benchname{} both read the image, so the correlations reported above rest on tasks that genuinely use the scan. Per-encoder floors are in Appendix~\ref{app:floor}.

\subsection{Generality to Merlin Dataset}
\label{sec:merlin}
\begin{figure}[t]
\centering
\includegraphics[width=\linewidth]{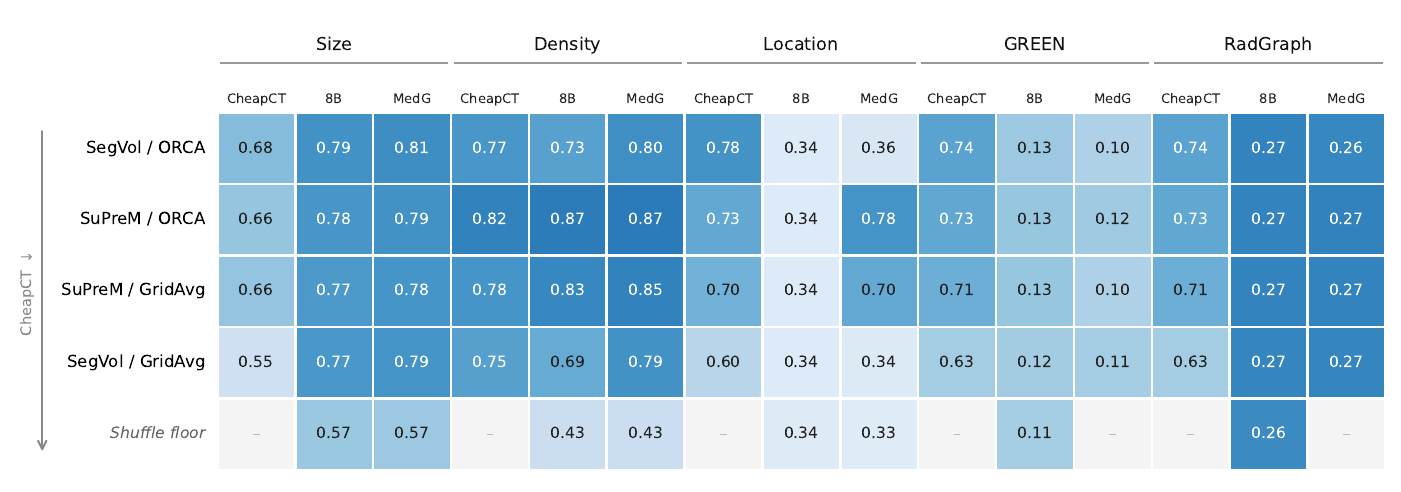}
\caption{Forecast agreement on Merlin. Rows are encoder cells sorted by CheapCT. Each capability gives the CheapCT score beside the fine-tuned score on Llama-8B and MedGemma, the shuffle floor beneath.}
\label{fig:merlin}
\end{figure}

We also validate the finding on Merlin, a contrast-enhanced abdominal corpus, with four encoder cells. That is too few for a stable correlation, so Figure~\ref{fig:merlin} shows the rank pattern. The cells are sorted by CheapCT. CheapCT tracks density on both backbones, and the gain over the shuffle floor stays wide. Location is the exception, where some cells barely train and CheapCT over-predicts them. Size already sits high on the abdomen for both backbones, so it does not separate. GREEN and RadGraph sit just above their shuffle floor in a narrow band, so their ranking says little.

\subsection{Cost}
On a single B200 GPU, a CheapCT probe trains in well under a minute, $14$ seconds on COLIPRI$_\mathrm{ORCA}$ and $49$ on ViSD+CT-CLIP. Fine-tuning and evaluating one language model on the same encoder takes hours. Report generation runs $10.7$ GPU-hours on COLIPRI$_\mathrm{ORCA}$ and $14.9$ on ViSD+CT-CLIP, \benchname{} $9.3$ and $6.6$. Either training costs hundreds to thousands of times more than the CheapCT probe that forecasts it.

\section{Discussion}
\noindent\textbf{Clinical evaluation.} CE-F1 scores a report against a fixed, limited set of abnormality labels. GREEN judges more of the report and reflects a broader clinical signal. CheapCT forecasts the narrow, label-based metric consistently. The broad judge is a noisier target and harder to forecast reliably. Making that forecast dependable is left to future work.

\noindent\textbf{The encoder ceiling.} The encoder is the high-leverage choice. Across the four backbones the absolute scores move, but the encoder ranking CheapCT recovers does not. What a 3D-CT vision language model can do is set largely by the encoder it starts from, not by the decoder placed on top. The encoder appears to bound what the system can reach, so building and improving encoders should return more than scaling the language model. CheapCT makes that selection cheap.

\noindent\textbf{From encoder to decoder.} CheapCT is a forecast, not a replacement for fine-tuning. It is a cheap first pass that narrows the encoder candidates, and its pick is still confirmed by one fine-tune. It also sees only the encoder. The language model has no such probe, so choosing the decoder still needs a full fine-tune, as our four-backbone comparison did. A natural next step carries the same cheap-proxy idea to the decoder. The goal is to forecast which language model or fine-tuning recipe wins without training each one, as AutoML forecasts trained accuracy from a cheap score.

\section{Conclusion}
\label{sec:conclusion}
We set out to test whether a cheap probe on a pretrained 3D-CT encoder forecasts the outcome of expensive language-model fine-tuning, and we built \benchname{} to answer it one capability at a time. Across five clinical capabilities a probe that trains in minutes predicts what full fine-tuning achieves, at orders of magnitude less compute, and it holds across language models and corpora. Spatial location is forecast by a position-aware readout. The forecast holds only where the fine-tuned model actually reads the image, a control we make explicit, so a practitioner can tell before paying whether to trust it. Together, \benchname{} and CheapCT reframe encoder choice for 3D-CT vision language models as a cheap measurement rather than an expensive search.

\newpage
\bibliographystyle{plainnat}
\bibliography{references}

\newpage
\appendix

\begin{center}
{\LARGE\bfseries Supplementary Material}\\[0.45em]
{\large\bfseries When Do Cheap Probes Predict Expensive Training? Probing 3D-CT Encoders for Text Generation}
\end{center}
\vspace{0.6em}

\section{MeasureVQA construction}
\label{app:vqa}

\subsection{Data extraction and filtering}
\benchname{} is built from two public 3D~CT datasets. CT-RATE~\citep{hamamci2026generalist} provides chest CT without contrast. Merlin~\citep{blankemeier2024merlin} provides predominantly portal venous abdominal CT. We use the training and validation splits of both, listed in Table~\ref{tab:app_datasets}, disjoint by volume and, in CT-RATE, by patient. Every measurement comes from a TotalSegmentator~\citep{wasserthal2023totalsegmentator} organ mask and the Hounsfield values within it. Size and location come from mask geometry, namely organ volumes, inter-organ ratios, anatomical diameters, and normalized positions. Density and texture come from the Hounsfield values, namely mean or median attenuation, fractions of high attenuation, intensity percentiles, and distribution shape. The abnormality labels come from the radiology reports and are asked as presence questions.

\begin{table}[!tbh]\centering\scriptsize
\setlength{\tabcolsep}{5pt}
\begin{tblr}{
    colspec = {l l l r c c c X[l]}, width=\linewidth, colsep=5pt, rowsep=1.6pt,
    row{1} = {font=\bfseries}, hline{1,4} = {0.08em}, hline{2} = {0.05em},
  }
  Dataset & Anatomy & Contrast & Patients & Volumes (tr/val) & Vols/pt. & Abnor. & Organs measured \\
  CT-RATE & chest & non-contrast & 21{,}304 & 47{,}149 / 3{,}039 & 2.4 & 18 & heart, lungs, aorta, IVC, vertebrae \\
  Merlin & abdomen & portal venous & 18{,}317 & 15{,}306 / 5{,}054 & 1.1 & 30 & liver, spleen, kidneys, aorta, vertebrae, paraspinal muscle \\
\end{tblr}
\caption{Source datasets used by \benchname{}. A CT-RATE patient contributes up to $32$ reconstructions.}
\label{tab:app_datasets}
\end{table}

We filter unreliable measurements per attribute using organ-level guards shared across the attributes that use each organ, listed in Table~\ref{tab:qc_guards}. An organ passes only when its mask is present, its volume lies in a plausible window, and, for the density measurements, its Hounsfield composition is right. When an organ fails, only the measurements that depend on it are withheld, and a volume is dropped only when every attribute of a capability is missing. The tertile cutpoints are the training-set $33$rd and $67$th percentiles, frozen and applied unchanged to validation, and a few attributes instead use clinical thresholds. The per-attribute definitions below give each measurement's exact extraction and any self-value clamp.

\begin{table}[!tbh]\centering\scriptsize
\setlength{\tabcolsep}{5pt}
\begin{tblr}{
    colspec = {l l l r l}, colsep=6pt, rowsep=1.4pt,
    row{1} = {font=\bfseries},
    cell{2}{1} = {r=5}{}, cell{7}{1} = {r=5}{},
    row{2-6} = {gray!12},
    hline{1,12} = {0.08em}, hline{2} = {0.05em}, hline{7} = {0.05em},
  }
  Corpus & Organ & Plausible volume & Min.\ mask (vox) & Composition guard \\
  CT-RATE & heart & 100--1600 mL & --- & --- \\
   & lung & 800--9000 mL & 100{,}000 & --- \\
   & aorta & 10--1000 mL & 20{,}000 & blood fraction $\ge 0.5$ \\
   & IVC & 2--400 mL & --- & --- \\
   & vertebra & --- & 20{,}000 & bone fraction $\ge 0.4$ (HU $> 150$) \\
  Merlin & spleen & $\ge 30$ mL & --- & --- \\
   & liver & $\ge 400$ mL & --- & --- \\
   & vertebra & $\ge 20$ mL & --- & --- \\
   & muscle & $\ge 40$ mL & --- & --- \\
   & kidney & both sides & $> 3000$/side & --- \\
\end{tblr}
\caption{Per-organ quality-control guards, shared by the attributes that use each organ. A measurement is withheld when its organ fails any guard.}
\label{tab:qc_guards}
\end{table}

\begin{table}[!tbh]\centering\scriptsize
\caption{CT-RATE (thorax) attributes: the value measured from each organ mask, the letter-question form it is discretized into, and the abnormality it captures. Disease is a report-derived reference axis, not an image measurement.}
\label{tab:attr_ctrate}
\begin{tblr}{
  colspec = {l l X[2.3,l] l X[1.9,l]},
  width = \linewidth,
  colsep = 4pt, rowsep = 1.6pt,
  cell{2}{1} = {r=5}{}, cell{7}{1} = {r=4}{}, cell{11}{1} = {r=3}{}, cell{14}{1} = {r=3}{},
  row{2-6} = {gray!12}, row{11-13} = {gray!12},
  hline{1,18} = {0.08em}, hline{2} = {0.05em}, hline{7,11,14,17} = {0.03em},
}
\textbf{Capability} & \textbf{Attribute} & \textbf{Measured value} & \textbf{Question} & \textbf{What it captures} \\
size & \texttt{sz\_heart\_lung} & $\log$(heart / lung volume) & tertile & cardiothoracic ratio, cardiomegaly \\
 & \texttt{sz\_aorta\_heart} & $\log$(aorta / heart volume) & tertile & aorto-cardiac size balance \\
 & \texttt{sz\_ivc\_aorta} & $\log$(IVC / aorta volume) & tertile & veno-arterial caliber \\
 & \texttt{aorta\_diameter\_mm} & maximum aortic diameter (mm) & clinical & aortic dilation, aneurysm \\
 & \texttt{heart\_width\_mm} & maximum cardiac width (mm) & tertile & heart size \\
density & \texttt{hu\_aorta\_calc} & fraction of aortic-wall voxels above a calcium HU cut & clinical & atherosclerotic calcium burden \\
 & \texttt{vert\_median} & median vertebral-body HU & tertile & bone mineral density \\
 & \texttt{lung\_mean} & mean lung density (HU) & tertile & diffuse lung disease, fluid \\
 & \texttt{hu\_lung\_haacon} & lung high-attenuation-area fraction & tertile & consolidation, ground glass \\
location & \texttt{lung\_LR\_logratio} & $\log$(left / right lung volume) & tertile & left--right lung balance \\
 & \texttt{heart\_x} & normalized left--right heart-centroid position & tertile & mediastinal laterality \\
 & \texttt{ivc\_z} & normalized cranio--caudal IVC-centroid position & tertile & supero-inferior landmark height \\
texture & \texttt{lung\_firstorder\_Kurtosis} & lung intensity-histogram kurtosis & tertile & high-density lung texture, fibrosis \\
 & \texttt{vert\_firstorder\_Kurtosis} & vertebral-body histogram kurtosis & tertile & trabecular bone texture \\
 & \texttt{lung\_Perc15} & lung 15th-percentile HU & clinical & emphysema, low attenuation \\
disease & 18 abnormalities & binary labels from radiology reports & presence & report-derived reference axis \\
\end{tblr}
\end{table}

\begin{table}[!tbh]\centering\scriptsize
\caption{Merlin (abdomen) attributes. Texture is not constructed for Merlin, and absolute organ HU is replaced by contrast-robust inter-organ density differences (liver minus spleen) because Merlin is contrast-enhanced.}
\label{tab:attr_merlin}
\begin{tblr}{
  colspec = {l l X[2.3,l] l X[1.9,l]},
  width = \linewidth,
  colsep = 4pt, rowsep = 1.6pt,
  cell{2}{1} = {r=3}{}, cell{5}{1} = {r=3}{}, cell{8}{1} = {r=2}{},
  row{2-4} = {gray!12}, row{8-9} = {gray!12},
  hline{1,11} = {0.08em}, hline{2} = {0.05em}, hline{5,8,10} = {0.03em},
}
\textbf{Capability} & \textbf{Attribute} & \textbf{Measured value} & \textbf{Question} & \textbf{What it captures} \\
size & \texttt{spleen\_ml} & splenic volume from the spleen mask & clinical & splenomegaly \\
 & \texttt{kidney\_ml} & combined left and right kidney volume & clinical & reduced total kidney volume \\
 & \texttt{aorta\_diameter\_mm} & maximum abdominal aortic diameter (mm) & tertile & abdominal aortic aneurysm \\
density & \texttt{muscle\_auto\_median} & median HU of the paraspinal (erector spinae) muscle & tertile & muscle fatty infiltration \\
 & \texttt{vert\_L1T12\_median} & median HU of the L1 and T12 vertebral bodies & tertile & bone mineral density \\
 & \texttt{liver\_spleen\_diff} & mean liver HU minus mean spleen HU & clinical & hepatic steatosis \\
location & \texttt{kidney\_z} & cranio--caudal renal-centroid position vs.\ lumbar span & tertile & kidney height \\
 & \texttt{kidney\_lr\_z\_asym} & left minus right renal height & tertile & renal height asymmetry \\
disease & 30 abnormalities & binary labels from radiology reports & presence & report-derived reference axis \\
\end{tblr}
\end{table}

\subsection{CT-RATE measurement attributes}
CT-RATE contributes fifteen measurements across size, density, location, and texture, listed in Table~\ref{tab:attr_ctrate}. Each entry below names the organ mask, gives the extracted quantity and its quality guard, and notes any clinical threshold that replaces the default tertile discretization.

\subsubsection{Size}
\noindent\textbf{\texttt{aorta\_diameter\_mm}} (aortic diameter). From the aorta mask of the thoracic TotalSegmentator segmentation. On each axial slice we take the largest circle that fits inside the aortic cross-section, its radius the maximum of the Euclidean distance transform of the slice mask under the physical in-plane spacing. The slice diameter is twice that radius. The attribute is the maximum of these slice diameters over the volume. This inscribed-circle, short-axis definition is deliberately robust to obliquely cut segments. Where the aortic arch crosses an axial plane lengthwise it forms a long thin region whose \emph{area} is large but whose inscribed circle is small, so the arch cannot inflate the diameter. The value is guarded to the plausible range $10$ to $80$~mm and split at the clinical threshold of $40$~mm into normal caliber ($<40$~mm) and dilated ($\ge 40$~mm), following the ACC/AHA aortic-disease guideline~\citep{isselbacher2022aorta}.

\noindent\textbf{\texttt{heart\_width\_mm}} (maximum transverse cardiac width). From the heart mask. On each axial slice we take the in-plane bounding-box extent of the heart, the larger of its left-right and anterior-posterior spans under the physical spacing, and the attribute is the maximum over slices. This is the numerator of the cardiothoracic ratio, the widest the cardiac silhouette becomes, which falls at the mid-ventricular level rather than among the great vessels. The value is guarded to $60$ to $250$~mm.

\noindent\textbf{\texttt{sz\_heart\_lung}} (heart-to-lung volume ratio). From the heart and lung masks. The attribute is the natural logarithm of the ratio of heart voxel count to total-lung voxel count, the five lung lobes summed. Because the voxel size cancels, this is the log of the volume ratio, a volumetric analogue of the cardiothoracic ratio, and larger values indicate a relatively larger heart. A ratio is scale free by construction, which is why it, rather than an absolute volume, is the size primitive on both corpora.

\noindent\textbf{\texttt{sz\_aorta\_heart}} (aorta-to-heart volume ratio). From the aorta and heart masks. The natural logarithm of the ratio of aortic to cardiac voxel count, again a log volume ratio, larger when the aorta is prominent relative to the heart.

\noindent\textbf{\texttt{sz\_ivc\_aorta}} (IVC-to-aorta volume ratio). From the inferior vena cava and aorta masks. The natural logarithm of the ratio of caval to aortic voxel count, a log volume ratio, larger when the inferior vena cava is prominent relative to the aorta.

\begin{figure}[!tbh]
\centering
\includegraphics[width=\linewidth]{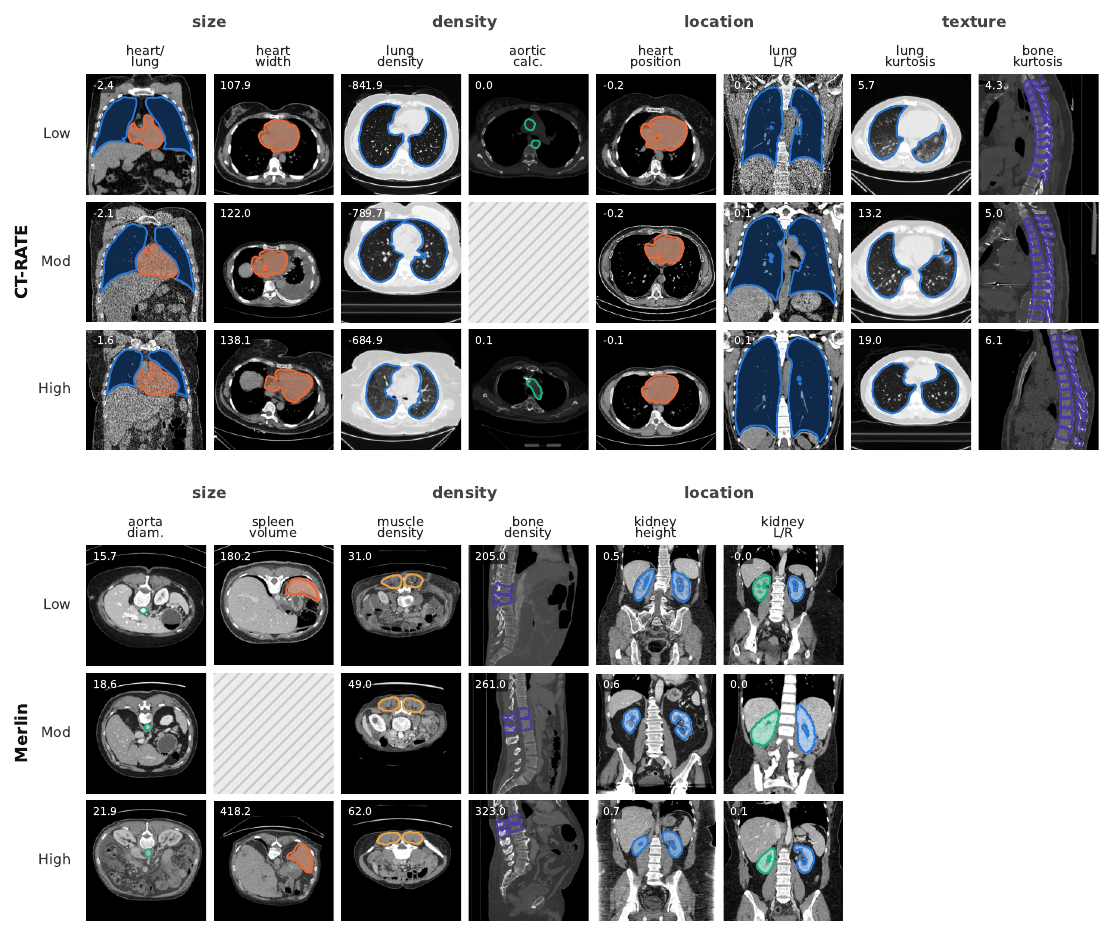}
\caption{\benchname{} dataset examples, CT-RATE (top) and Merlin (bottom). Columns are attributes by capability, rows the Low, Moderate, and High classes, each cell a CT slice with the organ mask and measured value. Binary attributes (aortic calcification, spleen volume) fill only the Low and High rows, the middle hatched.}
\label{fig:casestudy}
\end{figure}

\subsubsection{Density}
\noindent\textbf{\texttt{vert\_median}} (vertebral bone density). From the thoracic vertebral mask, the median Hounsfield value of the voxels inside the vertebral bodies. The median rather than the mean makes it robust to air pockets and metal, and a guard requires the mask to be bone dominated (bone fraction $\ge 0.4$) and of adequate size before a value is recorded. Low, moderate, and high tertiles run from relatively less to more dense bone. We deliberately do not apply a clinical osteoporosis threshold. Osteoporosis is defined on the L1 trabecular region of interest (around $100$~HU, excluding cortex and the posterior elements). We instead measure the whole-vertebra median, a different and higher quantity (around $261$~HU) on which that threshold never fires. Treating it as an osteoporosis test would be a clinical error, so the attribute falls back to relative tertiles~\citep{pickhardt2013osteoporosis, Jang2019Opportunistic}.

\noindent\textbf{\texttt{lung\_mean}} (mean lung density). From the lung mask, the five lobes, the mean Hounsfield value over all lung voxels. Low, moderate, and high tertiles run from the most aerated lungs, low mean attenuation toward emphysema or hyperinflation, to the densest, high attenuation toward effusion, consolidation, or fibrosis.

\noindent\textbf{\texttt{hu\_lung\_haacon}} (consolidation high-attenuation fraction). From the lung mask, the fraction of the lung occupied by consolidation. To keep the measure specific we first erode the lung by two voxels, dropping the pleural and mediastinal boundary where partial-volume averaging inflates attenuation, then take the voxels above $-250$~HU and morphologically open the result to remove the thin vessel tree. The attribute is the surviving high-attenuation voxel count over the eroded-lung voxel count.

\noindent\textbf{\texttt{hu\_aorta\_calc}} (aortic-wall calcification fraction). From the aorta mask, the fraction of aortic voxels that are calcified. Calcium is separated from contrast and noise by a per-volume adaptive threshold, the higher of a floor near $200$~HU and the blood-pool mean plus six standard deviations, keeping only connected components of at least ten voxels. The attribute is the kept calcified voxel count over the aortic voxel count. The distribution is zero inflated, since over a third of thoracic aortas carry no detectable calcification, so it ships as a present or absent question rather than as tertiles, with calcification present in $64\%$ of volumes.

\subsubsection{Location}
\noindent\textbf{\texttt{lung\_LR\_logratio}} (left-right lung volume balance). From the lung masks, the natural logarithm of the ratio of left to right lung voxel count. It is normally negative because the right lung, with three lobes, is larger than the two-lobed left. Values nearer zero indicate a more balanced or pathologically altered split.

\noindent\textbf{\texttt{heart\_x}} (cardiac lateral position). From the heart mask, the left-right position of the heart's centroid relative to the spine, expressed in half-lung-widths so that $0$ is the midline and the magnitude grows toward the lung edge (RAS convention, leftward negative). It captures how far off the midline, normally to the left, the cardiac silhouette sits. A centroid rather than a bounding box is used so a stray voxel cannot move it.

\noindent\textbf{\texttt{ivc\_z}} (inferior vena cava vertical position). From the inferior vena cava mask, the cranio-caudal position of the caval centroid as a fraction of the lung span, $0$ at the lung base and $1$ at the apex.

\subsubsection{Texture}
\noindent\textbf{\texttt{lung\_firstorder\_Kurtosis}} (lung density kurtosis). A first-order intensity feature, the kurtosis of the histogram of lung Hounsfield values, describing the tailedness of the attenuation distribution rather than any spatial pattern. Higher values mean the attenuation clusters sharply around its center with heavier tails.

\noindent\textbf{\texttt{vert\_firstorder\_Kurtosis}} (vertebral density kurtosis). The same first-order kurtosis over the vertebral Hounsfield distribution, a summary of how peaked and heavy tailed the bone-attenuation histogram is, reflecting the mixture of trabecular bone, cortex, and marrow.

\noindent\textbf{\texttt{lung\_Perc15}} (fifteenth-percentile lung density, emphysema). A densitometry index, the Hounsfield value below which the lowest $15\%$ of lung voxels fall. Lower, more negative values mean more low-attenuation, hyperinflated lung. It is split at the emphysema threshold of $-950$~HU into no significant emphysema ($\ge -950$~HU) and emphysema present ($< -950$~HU), a standard densitometric definition, distinct from the low-attenuation-area fraction below $-950$~HU~\citep{gevenois1995emphysema, lynch2015ct}.

\subsection{Merlin measurement attributes}
Merlin contributes eight measurements across size, density, and location, listed in Table~\ref{tab:attr_merlin}. The entries follow the format of the CT-RATE attributes above, and several reuse the same definitions.

\subsubsection{Size}
\noindent\textbf{\texttt{spleen\_ml}} (splenic volume). From the spleen mask, its voxel count times the voxel volume. Split at $314$~mL into normal splenic volume ($<314$~mL) and splenomegaly ($\ge 314$~mL), a CT-based splenomegaly threshold~\citep{prassopoulos1997splenic}.

\noindent\textbf{\texttt{kidney\_ml}} (total kidney volume). From the left and right kidney masks together, their combined volume, so a single functioning kidney reads as a reduced total. Split at $250$~mL into reduced ($<250$~mL) and normal ($\ge 250$~mL) total kidney volume.

\noindent\textbf{\texttt{aorta\_diameter\_mm}} (abdominal aortic diameter). The same inscribed-circle short-axis diameter as the CT-RATE aorta, but with the slice search restricted to the abdominal range, the levels at which a kidney is present. This restriction matters on Merlin. The mask runs from the thoracic aorta, normally the wider vessel near $35$~mm, down to the bifurcation, so an unrestricted maximum would read the chest rather than the abdominal aorta and manufacture a false aneurysm~\citep{chaikof2018abdominal}.

\subsubsection{Density}
\noindent\textbf{\texttt{muscle\_auto\_median}} (paraspinal muscle density). From the autochthon (erector spinae) muscle mask, the two bellies posterior to the vertebra, the median Hounsfield value. Lower values reflect fatty infiltration of muscle, higher values healthy muscle.

\noindent\textbf{\texttt{vert\_L1T12\_median}} (vertebral bone density). The median Hounsfield value of the L1 and T12 vertebral bodies, the whole-vertebra median as on CT-RATE rather than a trabecular region of interest, so the two corpora share one bone-density definition. As on CT-RATE, we do not read a whole-vertebra median as an osteoporosis test~\citep{pickhardt2013osteoporosis}.

\noindent\textbf{\texttt{liver\_spleen\_diff}} (liver-minus-spleen density, steatosis). The mean Hounsfield value of the liver minus that of the spleen. The difference form, rather than absolute liver attenuation, is used because it largely cancels the intravenous-contrast timing that varies across Merlin, since liver and spleen enhance together. A healthy liver is denser than the spleen, so a low or negative difference indicates fatty infiltration. We split at $-10$~HU into hepatic steatosis ($<-10$~HU) and no significant steatosis ($\ge -10$~HU)~\citep{doi:10.2214/AJR.09.2590}. The spleen is the weaker of the two segmentations, so spleen size is recorded as a quality column and volumes with a partial spleen are dropped.

\begin{figure}[!h]
\centering
\includegraphics[width=\linewidth]{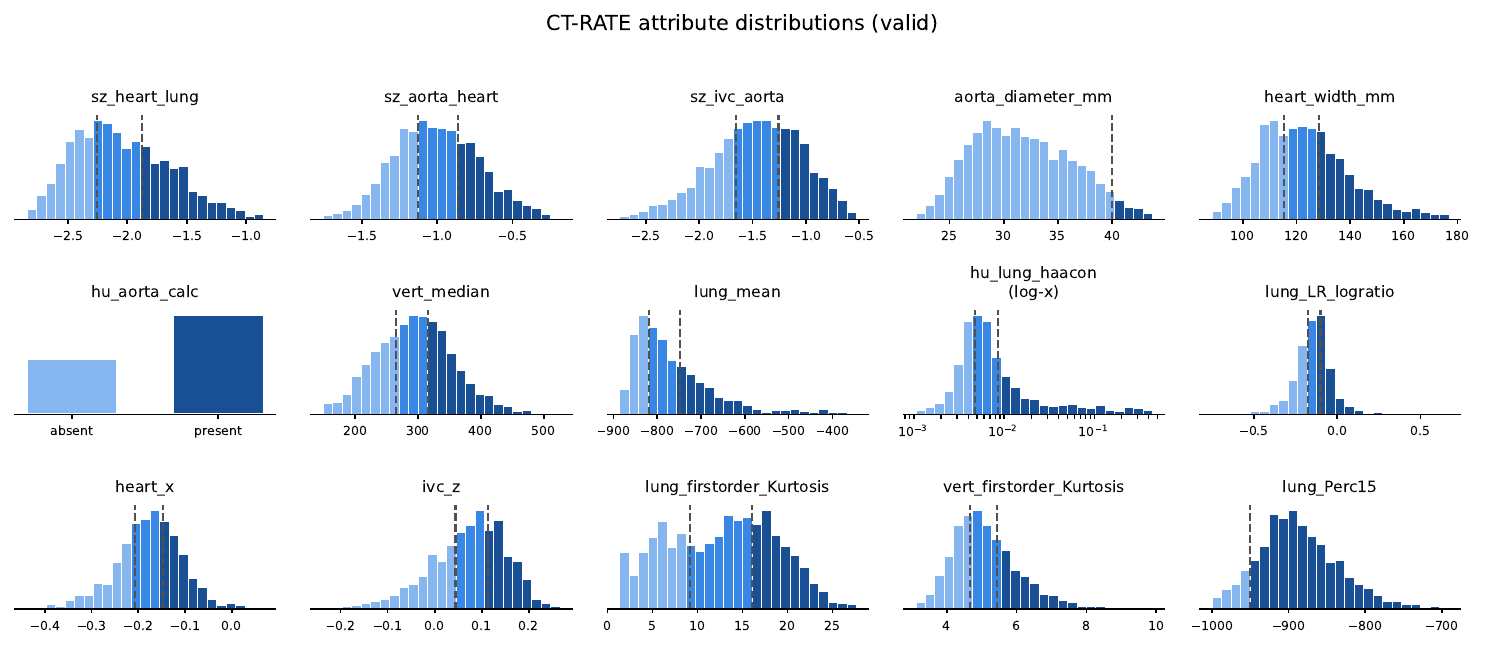}\\[2pt]
\includegraphics[width=\linewidth]{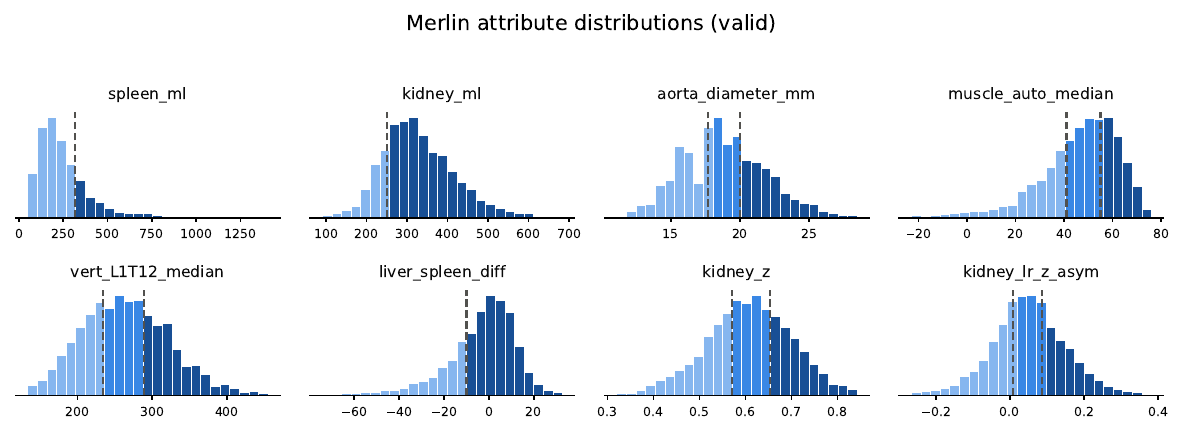}
\caption{Per-attribute value distributions on the validation split, bars colored by class, with the frozen cutpoints as dashed grey rules. Top: CT-RATE (15 attributes). Bottom: Merlin (8 attributes). Zero-inflated aortic-wall calcification is shown as present or absent, and right-skewed attributes use a logarithmic $x$-axis.}
\label{fig:pillar3_dist}
\end{figure}

\subsubsection{Location}
\noindent\textbf{\texttt{kidney\_z}} (kidney vertical position). From the kidney masks, the cranio-caudal position of the renal centroid measured against the lumbar vertebral span (L5 to T12), so it records where the kidneys sit relative to the spine rather than within the scan field. Higher values are more cranial. That it correlates only weakly with the scan's own vertical coverage ($r=0.10$) confirms it reflects anatomy, not field of view.

\noindent\textbf{\texttt{kidney\_lr\_z\_asym}} (left-right renal height asymmetry). The left-minus-right difference in cranio-caudal kidney position against the same lumbar span. The left kidney normally sits slightly higher, the right displaced downward by the liver, so typical values are small and positive. Larger values mark an unusually asymmetric pair.

\subsection{Questions and examples}
Every retained continuous measurement is converted into a question through a fixed template, with options shuffled reproducibly from the question identifier. Tertile questions contain three choices, clinical questions two, and report-derived abnormalities are rendered as binary presence questions. The continuous values are retained, and exact-value questions are generated from the same measurements.

Figure~\ref{fig:casestudy} shows one median slice per class for a broad selection of attributes. Size, volume, and density differences are directly visible. The heart fills more of the chest as the heart-to-lung ratio rises, the lungs brighten as their mean density rises, the spleen enlarges into splenomegaly, and the paraspinal muscle darkens with fatty infiltration. Position and texture are summaries of three-dimensional shape or intensity distributions and can be less apparent in a single slice.

\section{MeasureVQA audit}
\label{app:audit}

\subsection{Mask quality and measurement correctness}
\label{app:audit:mask}
\textbf{CT-RATE masks.} Of the $47{,}149$ official reconstructions, $47{,}146$ carry extracted metadata. Three were dropped at extraction for a mask or shape mismatch. Each organ the dataset uses is present in about $98\%$ of scans. A failure, an absent mask or a physiologically implausible organ volume, occurs in $1.5$ to $1.8\%$. The density guards go further, withholding a measurement when the mask is present but untrustworthy. The aorta guard passes on $95.1\%$ of volumes, so about one aortic-calcification measurement in twenty is withheld. Most are intravenous-contrast studies or mask bloat, where a bright lumen is read as calcium. The vertebra and lung guards pass on $98.3\%$ and $98.4\%$. The validation split shows the same rates.

\textbf{Merlin masks.} We read all $25{,}196$ abdominal masks directly. Segmentation is near-perfect for the organs the dataset uses. The liver, both kidneys, and the spine are each present in over $99.9\%$ of scans. Only ten masks fail quality control outright, every one for an absent or near-empty liver. A further $242$ volumes are soft-flagged for an outlying liver volume and kept. Every training and validation volume that carries one of our questions was measured.

\textbf{Repeatability.} Many CT-RATE scans are reconstructed more than once, which lets us check how stable a measurement is under a repeat acquisition. For each patient with at least two reconstructions we take the first two, compute the attribute on each, and correlate the paired values across patients with a Spearman rank correlation. Whole-vertebra median density reaches a test-retest of $0.969$, and the retained attributes clear a reliability threshold of $0.7$. This check also guided extraction. Eroding the lung mask and removing thin connected structures raised the test-retest of the consolidation high-attenuation fraction from $0.62$ to $0.86$.

\textbf{Answer re-derivation.} For every catalog question we re-derive the correct answer from the raw ground truth and the frozen cutpoints, then check it against the shipped answer letter. Over the fifteen CT-RATE and eight Merlin measurement targets on the validation split, every question matches, zero mismatches out of about $2{,}900$ to $3{,}000$ per target. The per-target counts dip exactly where a mask guard withholds a measurement, aortic calcification at $2{,}903$ and the texture targets at $2{,}566$. The withholding from the mask audit therefore carries through to the question set.

\subsection{Thresholds and balance}
\label{app:audit:clinical}
\noindent\textbf{Clinical thresholds.} Five attributes use literature clinical thresholds, thoracic aortic dilation at $40$ mm \citep{isselbacher2022aorta}, emphysema at the 15th percentile of lung density \citep{gevenois1995emphysema,lynch2015ct}, aortic wall calcification at $130$ HU \citep{AGATSTON1990827}, splenomegaly at $314$ mL \citep{prassopoulos1997splenic}, and hepatic steatosis at a liver-minus-spleen attenuation difference of $-10$ HU \citep{doi:10.2214/AJR.09.2590}. Total kidney volume uses a predefined study threshold of $250$ mL. The rest use training-split tertiles, because the measured quantity does not match a guideline definition. Vertebral density is computed over the whole vertebra rather than the L1 trabecular region used by osteoporosis thresholds \citep{pickhardt2013osteoporosis,Jang2019Opportunistic}. The heart-to-lung volume ratio differs from the width-based radiographic definition of cardiomegaly \citep{jcm10092016}. Abdominal aortic diameter uses tertiles because aneurysm cases are uncommon in Merlin, while CT-RATE keeps the $40$ mm threshold.

\textbf{Class balance.} Figure~\ref{fig:pillar3_dist} shows each attribute's distribution and where its cutpoints fall. Applying the frozen training cutpoints to validation places each tertile class between $0.30$ and $0.37$ on CT-RATE and $0.32$ to $0.35$ on Merlin. The clinical questions are deliberately less even. CT-RATE positive rates are $64\%$ for aortic wall calcification, $11\%$ for emphysema, and $4.7\%$ for aortic dilation. Merlin positive rates are $24\%$ for splenomegaly, $19\%$ for reduced total kidney volume, and $27\%$ for hepatic steatosis. Correct answer letters stay near uniform on both corpora, the tertile letters within $0.002$ of $1/3$ and the binary splits no more skewed than $0.503/0.497$.

\subsection{Shortcuts and capability structure}
\label{app:audit:shortcut}
\begin{figure}[!tbh]
\centering
\includegraphics[width=\linewidth]{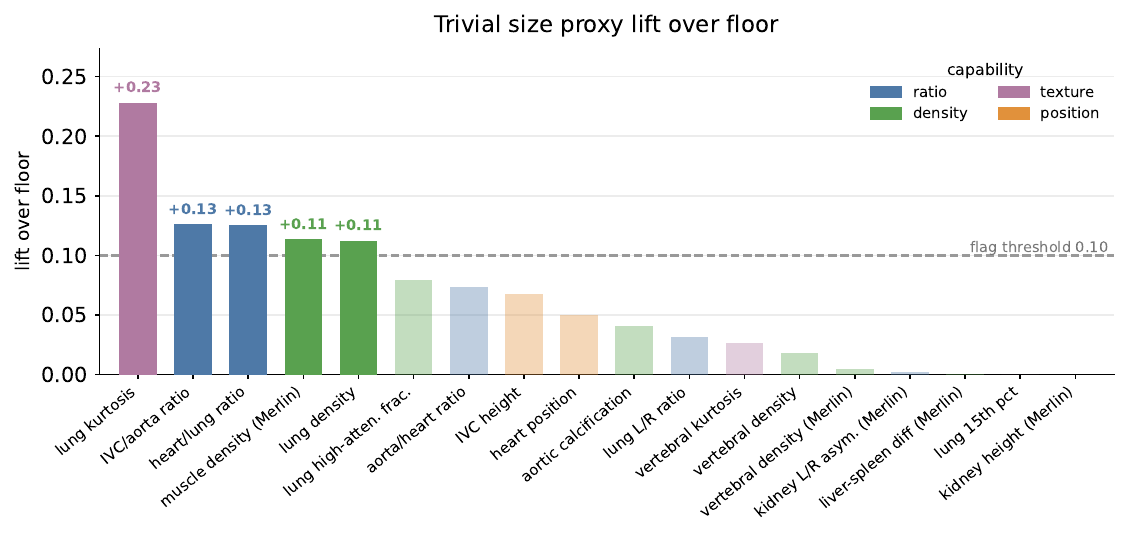}
\caption{Proxy accuracy lift over the honest floor $\max(\text{chance},\text{majority})$ for each invariance attribute, CT-RATE and Merlin pooled and colored by capability. Solid bars exceed the $0.10$ flag threshold, the five attributes with a modest size coupling. Size attributes are omitted as positive controls, where organ voxel count is a legitimate reading.}
\label{fig:trivial_proxy}
\end{figure}

\noindent\textbf{Trivial visual proxy.} For the per-capability results to mean anything, a density, texture, location, or ratio class must not be recoverable from gross organ size alone. We test the crudest size feature, the raw voxel count of the organ involved. For each attribute we bin volumes by that count on the training split, label each bin with its majority class, and apply the labels to validation. Recovery above the floor $\max(\text{chance},\ \text{majority class})$ would flag a size confound, as plotted in Figure~\ref{fig:trivial_proxy}. Size attributes are positive controls, since voxel count is a legitimate reading there.
Most invariance attributes stay at the floor. Five lift modestly, by at most $0.23$, the two volume ratios, lung density and its kurtosis, and Merlin muscle density. A large hyperinflated lung is at once bigger, less dense, and differently shaped in its histogram. The positive controls confirm the check has power, spleen voxel count tracking spleen volume at $\rho=0.84$. Even so the proxy never solves the flagged questions, its best accuracy $0.58$ against a floor of $0.35$, so most of each answer still needs the intended read. The residual coupling reflects genuine capability entanglement in the anatomy.

\medskip

\begin{figure}[!htb]
\centering
\includegraphics[width=\linewidth]{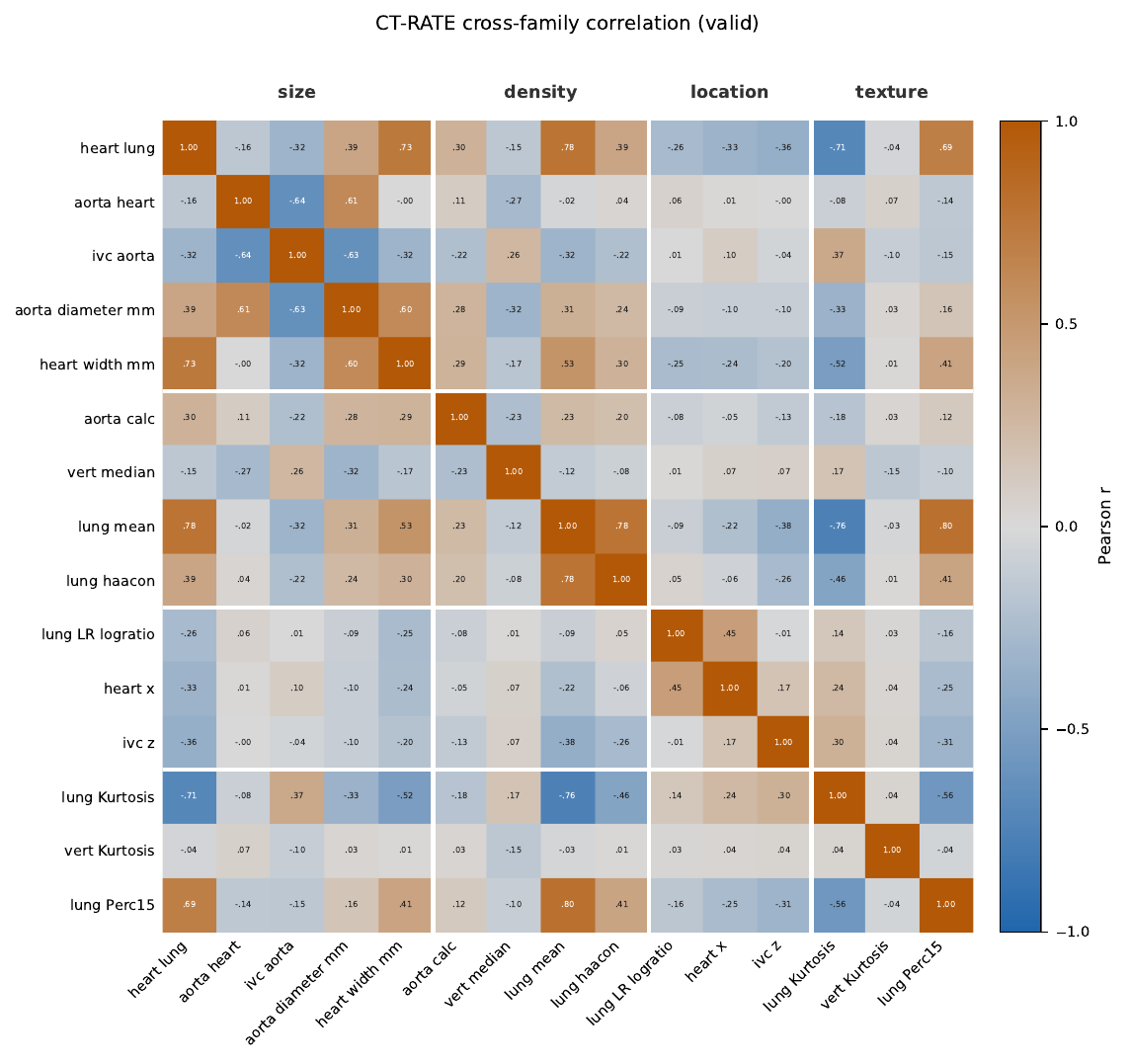}\\[3pt]
\includegraphics[width=0.62\linewidth]{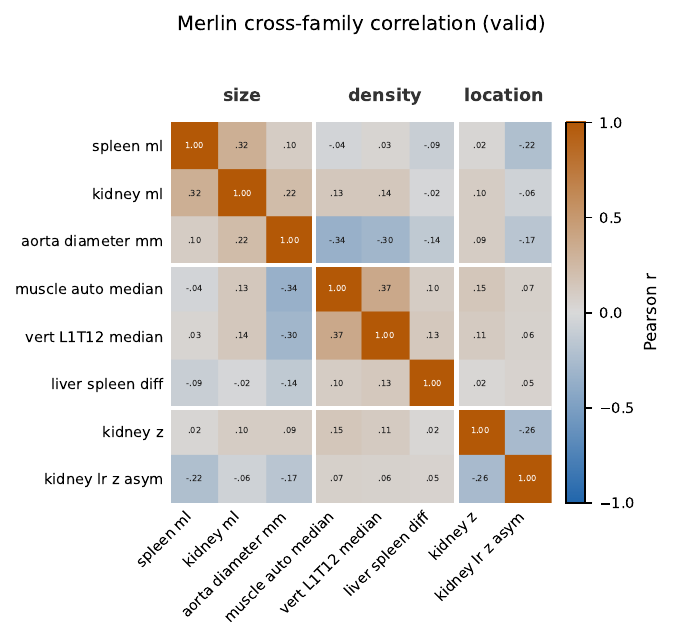}
\caption{Cross-capability attribute correlation on the validation split, Pearson $r$ blocked by capability. Top: CT-RATE (15 attributes). Bottom: Merlin (8 attributes).}
\label{fig:pillar5_crossfam}
\end{figure}
\noindent\textbf{Correlation structure.} Figure~\ref{fig:pillar5_crossfam} shows the full correlation matrix among measurement attributes. Within a capability, attributes are moderately correlated, the largest reaching $r=0.78$ in CT-RATE (mean lung density and the lung high-attenuation fraction) and $r=0.37$ in Merlin. Across capabilities, correlations are mostly smaller, the largest $0.80$ in CT-RATE and $0.34$ in Merlin. The CT-RATE exception is the lung, whose size, density, and texture share the pulmonary air-content axis, mean lung density correlating with the 15th percentile of lung density at $r=0.80$ and with lung density kurtosis at $r=-0.76$.

\begin{figure}[!tbh]
\centering
\includegraphics[width=\linewidth]{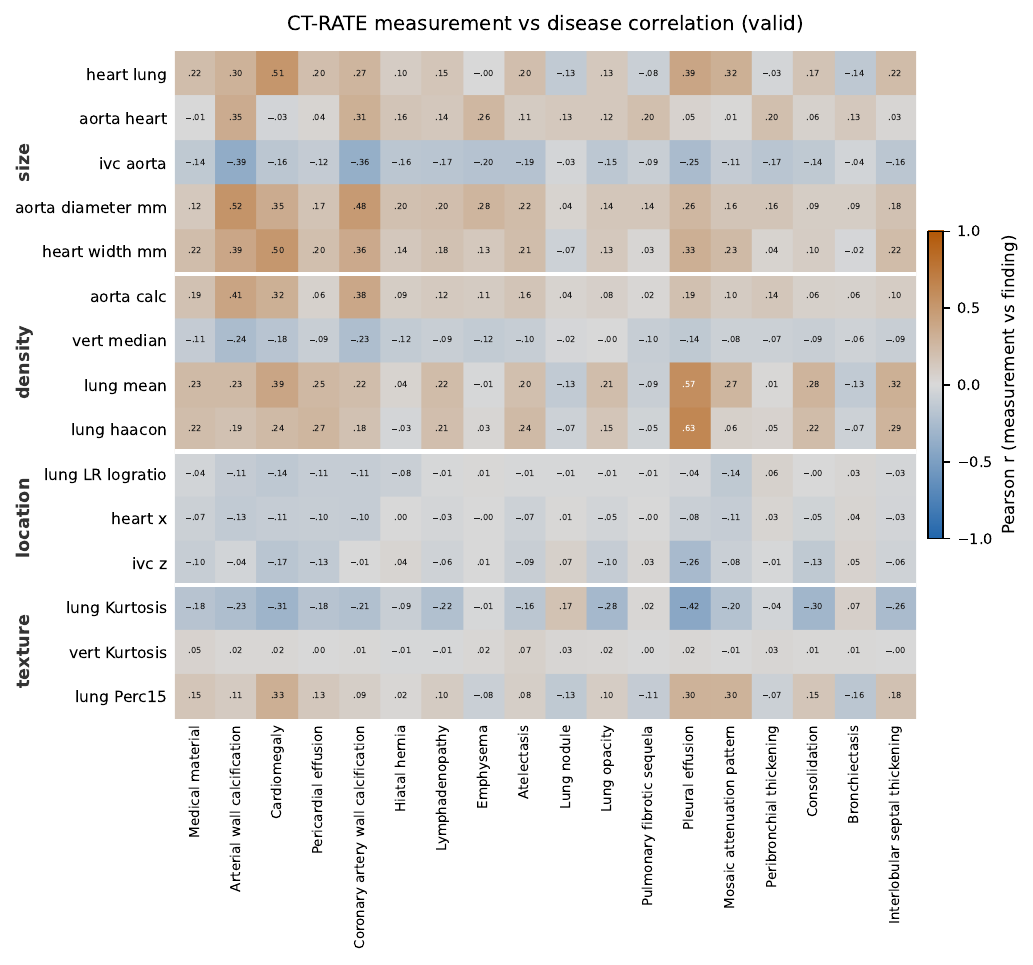}\\[3pt]
\includegraphics[width=\linewidth]{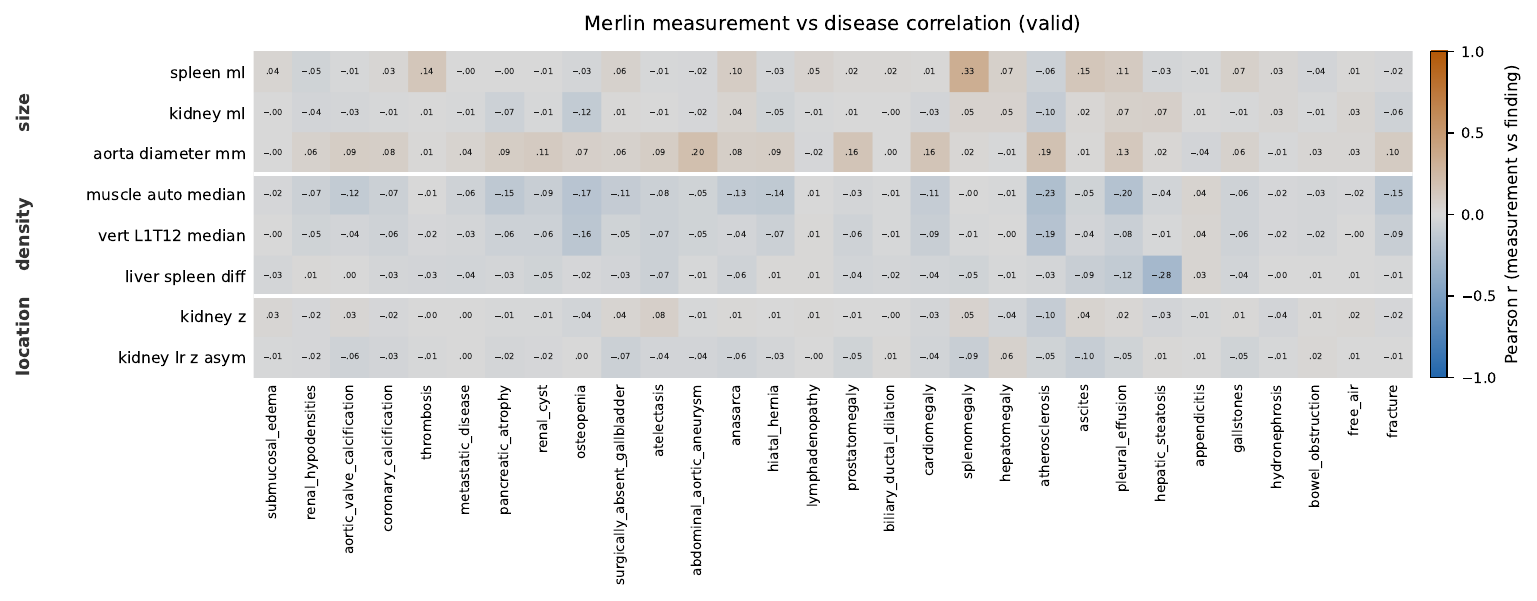}
\caption{Correlation between each continuous measurement (rows, blocked by capability) and each report-derived abnormality (columns), Pearson $r$ on the same diverging scale, validation split. Top: CT-RATE. Bottom: Merlin.}
\label{fig:pillar5_disease}
\end{figure}

\noindent\textbf{Relations to abnormalities.} Figure~\ref{fig:pillar5_disease} shows each measurement aligned with the abnormality it clinically implies. CT-RATE labels all eighteen abnormalities on every volume, with prevalence from $7\%$ to $45\%$. Merlin labels a sparse subset per report, missing not at random. An abnormality labeled on fewer volumes is positive more often among them, a correlation of $-0.43$, so Merlin disease numbers are reported over the available labels. In the extreme decile of a measurement the matching disease is enriched, the top decile of the heart-to-lung ratio holding $59\%$ cardiomegaly against $5\%$ in the cohort, with lifts of $1.4\times$ to $11\times$ across the seven CT-RATE pairs and $1.2\times$ to $70\times$ across the four Merlin pairs. Over the full distribution the correlation reaches $|r|=0.63$ in CT-RATE and $0.33$ in Merlin. Linear models using the disease labels explain at most $48\%$ of the variance of a CT-RATE measurement and $18\%$ of a Merlin measurement.

\medskip
\noindent\textbf{Limitations.} \benchname{} inherits errors from its data sources. Its measurements rely on TotalSegmentator organ masks, so local segmentation errors can affect the resulting values. Texture measurements are available only for CT-RATE, limiting them to thoracic CT. The shortcut analysis covers a coarse size proxy. \benchname{} evaluates predefined organ-level quantities, not lesion localization, open-ended clinical reasoning, or comprehensive diagnosis.

\begin{table}[!tbh]
\centering
\scriptsize
\begin{tblr}{
    colspec = {@{}lllrrl@{}}, colsep = 5pt, rowsep = 1.6pt,
    row{1} = {font=\bfseries},
    row{2,3,6,7,9} = {bg=black!8},
    hline{1,11} = {0.08em}, hline{2} = {0.05em},
  }
Encoder & Pretraining & Aggregation & Tokens & Dim & Ref \\
\SetCell[r=2]{l} COLIPRI & \SetCell[r=2]{l} Contrastive, with priors & ORCA & 216 & 792 & \SetCell[r=2]{l} \citep{wald2026comprehensivelanguageimagepretraining3d} \\
 & & GridAvg & 216 & 768 & \\
\SetCell[r=2]{l} CT-CLIP & \SetCell[r=2]{l} Contrastive & ORCA & 216 & 536 & \SetCell[r=2]{l} \citep{hamamci2026generalist} \\
 & & GridAvg & 216 & 512 & \\
\SetCell[r=2]{l} BTB3D & \SetCell[r=2]{l} Visual tokenizer & ORCA & 216 & 42 & \SetCell[r=2]{l} \citep{NEURIPS2025_c4861cc8} \\
 & & GridAvg & 216 & 18 & \\
ViSD & Organ-aware & native & 4 & 256 & \citep{11444916} \\
ViSD+CT-CLIP & Organ-aware, with global & native & 5 & 256 & \citep{liang2026embeddingbottleneckadaptiveretrievalaugmented} \\
fVLM & Anatomy-level & native & 4 & 256 & \citep{ICLR2025_3c3b120a} \\
\end{tblr}
\caption{The nine representation variants compared on CT-RATE.}
\label{tab:encoders}
\end{table}

\section{Experimental setup}
\label{app:setup}

\subsection{Encoders and representations}
We choose the encoders for diversity, spanning the pretraining strategies used for 3D CT today. Table~\ref{tab:encoders} details the nine representation variants. A dense encoder emits more tokens than the decoder should carry, so each is compressed to a fixed $216$-token budget by one of two aggregations. Organ-centroid pooling (ORCA)~\citep{liang2026orcaorgancentroidaggregationtrainingfree} groups tokens by organ. A plain grid average (GridAvg) averages within a fixed grid. Each dense encoder therefore contributes two variants. The organ-aware encoders already emit only a handful of tokens, so they need no aggregation and enter directly, one variant each. One of them combines the ViSD organ tokens with a single CT-CLIP global token, to isolate whether a global summary adds to the organ tokens.

\subsection{Implementation details}
\label{app:testbed}
\textbf{Metrics.} Report generation is scored primarily by CE-F1 from a RadBERT abnormality classifier~\citep{yan2022radbert}, with BLEU~\citep{papineni2002bleu}, ROUGE-L~\citep{lin-2004-rouge}, and METEOR~\citep{banerjee-lavie-2005-meteor} reported alongside. We also report GREEN~\citep{ostmeier-etal-2024-green}, which prompts a fine-tuned $7$B language model to compare a report against the reference and count the clinically significant matches and errors. We allow GREEN $1{,}024$ new output tokens, so its roughly $880$-token analysis is emitted in full rather than truncated. Our GREEN scores may therefore sit slightly above those reported under a tighter generation budget.

\textbf{Backbones.} We run four backbones, two Llama and two Gemma models. Llama-3.1-8B~\citep{https://doi.org/10.48550/arxiv.2407.21783} and Llama-3.3-70B span a change of scale, and Gemma-3-27B~\citep{gemmateam2025gemma3technicalreport} with its medically pretrained sibling MedGemma-27B~\citep{sellergren2026medgemmatechnicalreport} form a size-matched general and medical pair. Each backbone is an independent test of whether the probe forecasts its training. The decoder is adapted with LoRA~\citep{Hu2021LoRALA} on the attention and feed-forward projections (rank $128$, $\alpha=256$ for the $8$B and $27$B models, rank $64$, $\alpha=128$ for the $70$B, dropout $0.05$) while the base weights stay frozen.

\textbf{Training.} Training runs in two stages, warm-started one from the next. Stage one trains the projector alone at learning rate $5\times10^{-4}$, so the projector learns the embedding space before the model is touched. Stage two initialises a fresh LoRA and trains it together with the projector at learning rate $2\times10^{-5}$. Both stages use AdamW with a cosine schedule and $3\%$ warmup under a fixed seed, at an effective batch size of $16$ for every backbone. We evaluate at regular checkpoints and report each downstream metric at its best validation checkpoint, so the MeasureVQA accuracy and each report-generation metric are read at the step that maximizes that metric.

\textbf{Inference.} Report generation is run with vLLM under greedy decoding matched to training, at most $512$ new tokens and repetition penalty $1.3$, with one shared inference path across encoders. Matching the repetition penalty is what aligns vLLM with a standard batch-one decode. Even then vLLM shifts CE-F1 by about $0.01$ from floating-point kernel order, and it leaves the encoder ranking unchanged.

\textbf{Hardware.} Each language-model run uses a single NVIDIA B200 GPU ($192$\,GB), the 70B backbone included under LoRA and gradient checkpointing.

\begin{table}[!tbh]\centering\scriptsize
\begin{tblr}{
    colspec = {@{}l X[l] l c c l@{}}, colsep = 5pt, rowsep = 1.4pt,
    row{1} = {font=\bfseries},
    hline{1,10} = {0.08em}, hline{2} = {0.05em},
  }
Readout & Pooling mechanism & Params & Attn. & Out-dim & Capacity \\
Token mean & average over the tokens & --- &  & $C$ & weak \\
Max & per-channel maximum over tokens & --- &  & $C$ & weak \\
GeM & generalized-mean pool & learnable $p$ &  & $C$ & weak \\
Top-$k$ & mean of the top $k$ tokens per channel & $k{=}32$ &  & $C$ & weak \\
Gated attention & one attention head weights and sums the tokens & 1 head & \checkmark & $C$ & medium \\
Multi-head & multi-head gated attention & 4 heads & \checkmark & $4C$ & medium \\
\textbf{Position-aware} & multi-head attention with a per-slot embedding & 4 heads & \checkmark & $4C$ & medium \\
Dual-pool & transformer, then a global-and-local pool & $L{=}2$ & \checkmark & $2r$ & high \\
\end{tblr}
\caption{The readout structures compared in the robustness sweep, all sharing the same two-layer output head and one training protocol. The headline readout (\textbf{position-aware}) is a multi-head attention pool with a per-slot position embedding, and $C$ is the encoder's token dimension. Reading power (transformer depth $L$) and input normalization are varied separately as ablations.}
\label{tab:readouts}
\end{table}

\subsection{Probe readout and training}
\label{app:probe}
\textbf{Readout.} A CT encoder variant presents each volume as visual tokens, and a probe reads them with a small, cheap network that pools the tokens into one vector and predicts the capability. We vary the readout across a family, listed in Table~\ref{tab:readouts}, and report the strongest all-round readout as the headline, a position-aware multi-head attention pool built on gated attention~\citep{pmlr-v80-ilse18a}. It is best or near-best on every capability, and because it keeps token positions it is the one readout that can forecast the spatial location capability. They all share the same downstream parts, a two-layer head on the pooled vector and a non-affine layer-norm on the input tokens.

\begin{figure}[!tbh]
\centering
\includegraphics[width=0.9\linewidth]{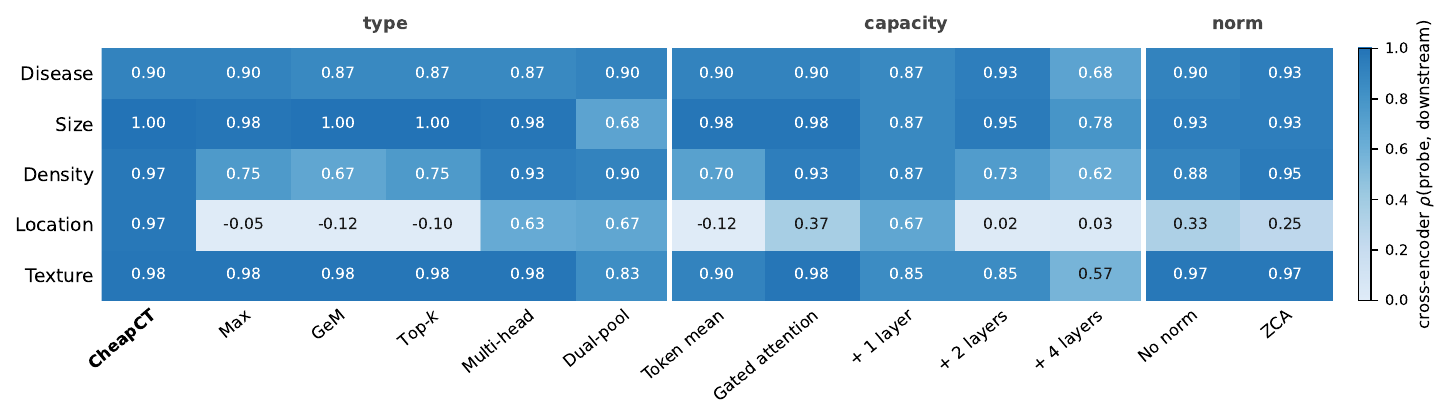}
\caption{Readout robustness, full sweep. Cross-encoder forecast agreement $\rho$ for every readout against every capability, on the Llama-3.1-8B backbone.}
\label{fig:sensitivity_ablation}
\end{figure}

\textbf{Training.} Each capability is predicted jointly by a single probe, disease as a binary classification over the CT-RATE abnormalities with a class-balanced cross-entropy, and each measurement capability as a regression on the z-standardized mask-derived value with a masked mean-squared error that ignores absent values. Every probe is trained with AdamW at learning rate $10^{-3}$, batch size $64$, for at most $50$ epochs with early stopping, the best epoch chosen by validation AUROC for disease and $R^2$ for the measurements. We run three seeds and report the mean and its spread.

\begin{table}[!tbh]
\centering
\scriptsize
\begin{tabular}{@{}lcccccc@{}}
\toprule
& \multicolumn{3}{c}{CE-F1} & \multicolumn{3}{c}{GREEN} \\
\cmidrule(lr){2-4}\cmidrule(lr){5-7}
Encoder & Real & Noise & Shuffle & Real & Noise & Shuffle \\
\midrule
BTB3D$_\mathrm{GridAvg}$   & $0.31$ & $0.20$ & $0.31$ & $0.25$ & $0.17$ & $0.25$ \\
BTB3D$_\mathrm{ORCA}$   & $0.34$ & $0.28$ & $0.27$ & $0.30$ & $0.09$ & $0.29$ \\
COLIPRI$_\mathrm{GridAvg}$ & $0.49$ & $0.23$ & $0.27$ & $0.33$ & $0.31$ & $0.30$ \\
COLIPRI$_\mathrm{ORCA}$ & $0.49$ & $0.19$ & $0.27$ & $0.35$ & $0.12$ & $0.31$ \\
CT-CLIP$_\mathrm{GridAvg}$ & $0.34$ & $0.29$ & $0.26$ & $0.31$ & $0.22$ & $0.29$ \\
CT-CLIP$_\mathrm{ORCA}$ & $0.37$ & $0.26$ & $0.28$ & $0.30$ & $0.17$ & $0.28$ \\
fVLM                    & $0.44$ & $0.18$ & $0.28$ & $0.32$ & $0.02$ & $0.30$ \\
ViSD-Boost              & $0.47$ & $0.10$ & $0.27$ & $0.31$ & $0.01$ & $0.28$ \\
ViSD+CT-CLIP            & $0.49$ & $0.12$ & $0.28$ & $0.32$ & $0.00$ & $0.30$ \\
\bottomrule
\end{tabular}
\caption{Report-generation floor per encoder on CT-RATE with Llama-3.1-8B. Clinical-efficacy F1 and GREEN, each with the real image and under the noise and shuffle floors.}
\label{tab:floor}
\end{table}

\section{Results}
\label{app:results}
This section gathers the evidence that the finding holds. We fix no threshold in advance. We read the finding off the numbers and ask whether it survives every check we can apply.

\subsection{Credibility of the finding}
\label{app:credibility}
A high cross-encoder correlation could arise from something other than a genuine, readable capability. We list those alternatives and show the finding outlasts each, so the forecastability is a property of the representations rather than an artifact of the readout we chose, a downstream task that ignores the image, or an organ-size shortcut.

\textbf{Not one lucky readout.} The verdict could be an accident of the particular pooling head. We sweep the readout over a band of thirteen choices spanning its capacity, its pooling type, and the input normalization, shown in Figure~\ref{fig:sensitivity_ablation}. For size, density, texture, and disease the verdict is the same for every readout, so it is a property of the encoder's tokens rather than of the readout. Location is the exception. It is readable only by a position-aware pool, near $\rho=0$ for the permutation-invariant pools and $\rho=0.96$ for the position-aware one, because location is a spatial relation the pooling can keep or discard. We therefore adopt the position-aware pool as the headline, the strongest all-round probe in the family and the one that supplies the spatial information location needs.

\textbf{The downstream task reads the image.} A capability can be forecast only if the downstream score itself depends on the image. Each pairing is checked against the score the task keeps with the image withheld (Section~\ref{sec:floor}). Report generation clears this blind floor by a wide margin, CE-F1 $0.42$ against $0.20$ under noise and $0.28$ under shuffle, and \benchname{} likewise, $0.68$ against $0.42$. We read forecastability only where the downstream task earns its score from the scan.

\textbf{Not a size shortcut.} A probe could score well by reading overall organ size in place of the intended attribute. The invariance attributes, built so that size carries no information about them, are held to their analytic floor, and the probe stays near it (Appendix~\ref{app:audit:shortcut}). A few small couplings remain, which we disclose rather than hide. The probe is reading the attribute, not its size.

\subsection{Floor control per encoder}
\label{app:floor}
Table~\ref{tab:floor} gives the report-generation floor per encoder, for CE-F1 and GREEN. The shuffle floors are close across the encoders, but the real score is not, so the image gain over the shuffle floor grows with encoder quality. On CE-F1 the strongest encoders clear it by twenty to thirty points. The weakest sit almost on it, their report barely beating a mismatched scan, exactly as expected of an encoder that carries little. GREEN moves the same way but compresses the range, so its gain over the shuffle floor is only a few points. Its noise floor is more scattered, since random features can still emit fluent boilerplate that the metric partly rewards. The measurement floors in Table~\ref{tab:main_results}, by contrast, vary little across encoders.

\begin{table*}[!tbh]
\centering
\scriptsize
\setlength{\tabcolsep}{2.5pt}
\begin{tabular}{@{}l ccc ccc ccc ccc@{}}
\toprule
& \multicolumn{3}{c}{Llama-3.1-8B} & \multicolumn{3}{c}{Llama-3.3-70B} & \multicolumn{3}{c}{Gemma-3-27B} & \multicolumn{3}{c}{MedGemma-27B} \\
\cmidrule(lr){2-4}\cmidrule(lr){5-7}\cmidrule(lr){8-10}\cmidrule(lr){11-13}
Encoder & CE-F1 & GREEN & RG & CE-F1 & GREEN & RG & CE-F1 & GREEN & RG & CE-F1 & GREEN & RG \\
\midrule
BTB3D$_\mathrm{GridAvg}$ & $0.31$ & $0.25$ & $0.17$ & $0.30$ & $0.29$ & $0.18$ & $0.30$ & $0.17$ & $0.13$ & $0.30$ & $0.14$ & $0.12$ \\
BTB3D$_\mathrm{ORCA}$ & $0.34$ & $0.30$ & $0.21$ & $0.31$ & $0.32$ & $0.20$ & $0.32$ & $0.15$ & $0.12$ & $0.33$ & $0.14$ & $0.12$ \\
COLIPRI$_\mathrm{GridAvg}$ & $0.49$ & $0.33$ & $0.23$ & $0.48$ & $0.36$ & $0.21$ & $0.44$ & $0.21$ & $0.11$ & $0.45$ & $0.23$ & $0.16$ \\
COLIPRI$_\mathrm{ORCA}$ & $0.49$ & $0.35$ & $0.22$ & $0.46$ & $0.38$ & $0.20$ & $0.45$ & $0.27$ & $0.17$ & $0.45$ & $0.27$ & $0.18$ \\
CT-CLIP$_\mathrm{GridAvg}$ & $0.34$ & $0.31$ & $0.21$ & $0.34$ & $0.34$ & $0.21$ & $0.31$ & $0.23$ & $0.15$ & $0.32$ & $0.16$ & $0.12$ \\
CT-CLIP$_\mathrm{ORCA}$ & $0.37$ & $0.30$ & $0.21$ & $0.35$ & $0.34$ & $0.21$ & $0.30$ & $0.17$ & $0.13$ & $0.33$ & $0.19$ & $0.14$ \\
fVLM & $0.44$ & $0.32$ & $0.22$ & $0.44$ & $0.37$ & $0.21$ & $0.34$ & $0.16$ & $0.14$ & $0.34$ & $0.13$ & $0.12$ \\
ViSD-Boost & $0.47$ & $0.31$ & $0.21$ & $0.48$ & $0.37$ & $0.22$ & $0.39$ & $0.19$ & $0.15$ & $0.40$ & $0.17$ & $0.13$ \\
ViSD+CT-CLIP & $0.49$ & $0.32$ & $0.22$ & $0.47$ & $0.37$ & $0.21$ & $0.34$ & $0.17$ & $0.13$ & $0.39$ & $0.21$ & $0.15$ \\
\midrule
Spearman $\rho$ & $0.90$ & $0.97$ & $0.92$ & $0.83$ & $0.90$ & $0.43$ & $0.90$ & $0.50$ & $0.30$ & $0.93$ & $0.77$ & $0.73$ \\
Pearson $r$ & $0.99$ & $0.83$ & $0.79$ & $0.98$ & $0.93$ & $0.67$ & $0.83$ & $0.44$ & $0.31$ & $0.91$ & $0.67$ & $0.66$ \\
Weighted Kendall $\tau$ & $0.73$ & $0.94$ & $0.77$ & $0.49$ & $0.82$ & $0.14$ & $0.88$ & $0.57$ & $0.37$ & $0.90$ & $0.82$ & $0.80$ \\
\bottomrule
\end{tabular}
\caption{Report-generation scores per encoder on CT-RATE for each backbone. RG is RadGraph-F1.}
\label{tab:backbone_rg}
\end{table*}

\begin{table*}[!tbh]
\centering
\scriptsize
\setlength{\tabcolsep}{4pt}
\begin{tabular}{@{}l cccc cccc@{}}
\toprule
& \multicolumn{4}{c}{Llama-3.1-8B} & \multicolumn{4}{c}{MedGemma-27B} \\
\cmidrule(lr){2-5}\cmidrule(lr){6-9}
Encoder & Size & Density & Location & Texture & Size & Density & Location & Texture \\
\midrule
BTB3D$_\mathrm{GridAvg}$ & $0.66$ & $0.65$ & $0.55$ & $0.69$ & $0.64$ & $0.66$ & $0.53$ & $0.68$ \\
BTB3D$_\mathrm{ORCA}$ & $0.76$ & $0.70$ & $0.61$ & $0.72$ & $0.76$ & $0.72$ & $0.67$ & $0.71$ \\
COLIPRI$_\mathrm{GridAvg}$ & $0.77$ & $0.81$ & $0.67$ & $0.79$ & $0.77$ & $0.78$ & $0.72$ & $0.80$ \\
COLIPRI$_\mathrm{ORCA}$ & $0.77$ & $0.84$ & $0.72$ & $0.84$ & $0.77$ & $0.82$ & $0.65$ & $0.81$ \\
CT-CLIP$_\mathrm{GridAvg}$ & $0.76$ & $0.73$ & $0.68$ & $0.75$ & $0.75$ & $0.71$ & $0.67$ & $0.74$ \\
CT-CLIP$_\mathrm{ORCA}$ & $0.77$ & $0.71$ & $0.67$ & $0.75$ & $0.71$ & $0.67$ & $0.61$ & $0.72$ \\
fVLM & $0.77$ & $0.72$ & $0.60$ & $0.74$ & $0.77$ & $0.72$ & $0.57$ & $0.74$ \\
ViSD-Boost & $0.81$ & $0.78$ & $0.55$ & $0.77$ & $0.81$ & $0.75$ & $0.54$ & $0.76$ \\
ViSD+CT-CLIP & $0.82$ & $0.81$ & $0.57$ & $0.79$ & $0.82$ & $0.77$ & $0.52$ & $0.77$ \\
\midrule
Spearman $\rho$        & $0.93$ & $0.93$ & $0.97$ & $0.93$ & $0.93$ & $0.93$ & $0.70$ & $0.98$ \\
Pearson $r$            & $0.96$ & $0.91$ & $0.92$ & $0.93$ & $0.94$ & $0.89$ & $0.78$ & $0.93$ \\
Weighted Kendall $\tau$ & $0.89$ & $0.84$ & $0.83$ & $0.87$ & $0.91$ & $0.90$ & $0.50$ & $0.97$ \\
\bottomrule
\end{tabular}
\caption{\benchname{} tertile accuracy per encoder on CT-RATE, for the two backbones we fine-tune for it.}
\label{tab:backbone_mvqa}
\end{table*}

\subsection{Downstream scores per backbone}
\label{app:backbone}
We give the scores behind the backbone check. Table~\ref{tab:backbone_rg} lists the report generation CE-F1, GREEN, and RadGraph per encoder on each of the four backbones, and Table~\ref{tab:backbone_mvqa} the \benchname{} accuracy per capability on the two backbones we fine-tune for it. CE-F1 ranks the encoders consistently on every backbone, and GREEN and RadGraph track it too, less consistently. RadGraph-F1 gives every encoder a similar low score, roughly $0.11$ to $0.23$, far tighter than the CE-F1 range of $0.30$ to $0.49$. The gaps between encoders are small next to the metric noise, so their rank order shifts from backbone to backbone. The CheapCT probe score does not depend on the backbone. Gemma-3-27B is a general instruct model, and its reports carry conversational preambles and non-radiology formatting that the GREEN judge scores unstably. We regenerate them with a style instruction that asks for the report alone, which lifts the Spearman $\rho$ between CheapCT and GREEN from $0.20$ to $0.50$. The residual weakness is the judge out of distribution on Gemma, so we report the style-forced GREEN and rely on CE-F1.

\subsection{Training curve}
\label{app:training}
\begin{figure}[!tbh]
\centering
\includegraphics[width=\linewidth]{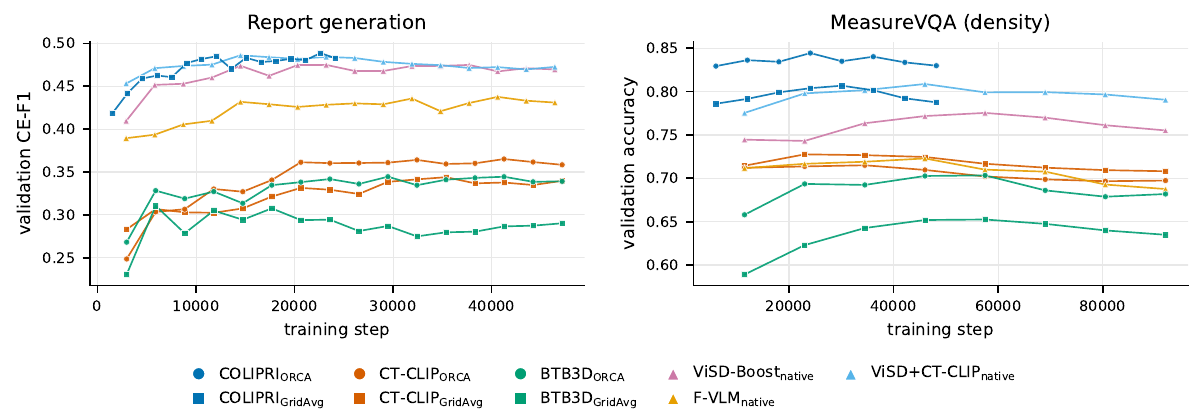}
\caption{Fine-tuning curves per encoder on Llama-3.1-8B. The left panel is report generation CE-F1. The right panel is MeasureVQA size. Colour gives the encoder. Marker gives the token aggregation.}
\label{fig:training_curve}
\end{figure}

Figure~\ref{fig:training_curve} tracks the validation CE-F1 through report-generation fine-tuning on Llama-3.1-8B, one curve per encoder. Every run converges, and the encoders separate early and keep their order, the stronger settling at a higher CE-F1 and the weaker plateauing low. The ranking CheapCT forecasts is a stable property of the representation, visible long before the run ends.

\end{document}